\documentclass[10pt,twocolumn,letterpaper]{article}

\usepackage{graphicx}
\usepackage{amsmath}
\usepackage{amssymb}
\usepackage{booktabs}
\usepackage{diagbox}
\usepackage{multirow}
\usepackage{makecell}
\usepackage{tabularx}
\usepackage{marvosym}
\usepackage{verbatim}
\usepackage{colortbl}
\usepackage{float}
\usepackage{color}
\usepackage[accsupp]{axessibility}  % Improves PDF readability for those with disabilities.
\usepackage{amsmath,amsfonts,bm}

\DeclareMathAlphabet{\mathsfit}{\encodingdefault}{\sfdefault}{m}{sl}
\SetMathAlphabet{\mathsfit}{bold}{\encodingdefault}{\sfdefault}{bx}{n}

\def\gF{{\mathcal{F}}}

\def\gL{{\mathcal{L}}}

\newcommand{\R}{\mathbb{R}}

\usepackage[pagenumbers]{cvpr} % To force page numbers, e.g. for an arXiv version

\definecolor{cvprblue}{rgb}{0.21,0.49,0.74}
\usepackage[pagebackref,breaklinks,colorlinks,allcolors=cvprblue]{hyperref}

\def\paperID{8964} % *** Enter the Paper ID here
\def\confName{CVPR}
\def\confYear{2025}

\title{Advancing Generalizable Tumor Segmentation with Anomaly-Aware Open-Vocabulary Attention Maps and Frozen Foundation Diffusion Models}

\author{Yankai Jiang\textsuperscript{1}, \quad Peng Zhang\textsuperscript{2}, \quad Donglin Yang\textsuperscript{3}, \quad Yuan Tian\textsuperscript{1}, \\ Hai Lin\textsuperscript{2},  \quad  Xiaosong Wang\textsuperscript{1}\\
\textsuperscript{1}Shanghai AI Laboratory \quad
\textsuperscript{2}Zhejiang University \quad \textsuperscript{3}The University of British Columbia \\
\tt\small{jiangyankai@pjlab.org.cn, lin@cad.zju.edu.cn, wangxiaosong@pjlab.org.cn} \\
}

\begin{document}

\maketitle

\begin{abstract}
We explore Generalizable Tumor Segmentation, aiming to train a single model for zero-shot tumor segmentation across diverse anatomical regions. Existing methods face limitations related to segmentation quality, scalability, and the range of applicable imaging modalities. In this paper, we uncover the potential of the internal representations within frozen medical foundation diffusion models as highly efficient zero-shot learners for tumor segmentation by introducing a novel framework named \textbf{DiffuGTS}. \textbf{DiffuGTS} creates anomaly-aware open-vocabulary attention maps based on text prompts to enable generalizable anomaly segmentation without being restricted by a predefined training category list. To further improve and refine anomaly segmentation masks, \textbf{DiffuGTS} leverages the diffusion model, transforming pathological regions into high-quality pseudo-healthy counterparts through latent space inpainting, and applies a novel pixel-level and feature-level residual learning approach, resulting in segmentation masks with significantly enhanced quality and generalization. Comprehensive experiments on four datasets and seven tumor categories demonstrate the superior performance of our method, surpassing current state-of-the-art models across multiple zero-shot settings. Codes are available at \url{https://github.com/Yankai96/DiffuGTS}.
%where the model encounters unseen medical modalities and anatomical regions during training.
%Recently, there has been a growing interest in expanding the application of generative models from generation tasks to semantic segmentation. These approaches utilize generative models either for generating annotated data or extracting features to facilitate semantic segmentation. This typically involves generating a considerable amount of synthetic data or requiring additional mask annotations.
\end{abstract}    
\section{Introduction}
\label{sec:intro}
Generalizable tumor segmentation (GTS) represents a fundamental challenge within medical image analysis~\cite{antonelli2022medical,liu2023clip,chen2024towards,jiang2024zept,wu2024freetumor}, stemming from both the diversity of tumor types and the variability across imaging modalities.
Current AI models for multi-tumor segmentation~\cite{liu2023clip,chen2023cancerunit,ye2023uniseg,zhao2023one,gao2024training} rely heavily on comprehensively annotated training data, limiting their ability to generalize beyond a restricted set of categories. This makes it challenging to address unseen diseases in clinical scenarios where the available training data may not adequately represent the diversity of real-world cases.

\begin{figure*}
  \centering
  \includegraphics[width=\linewidth]{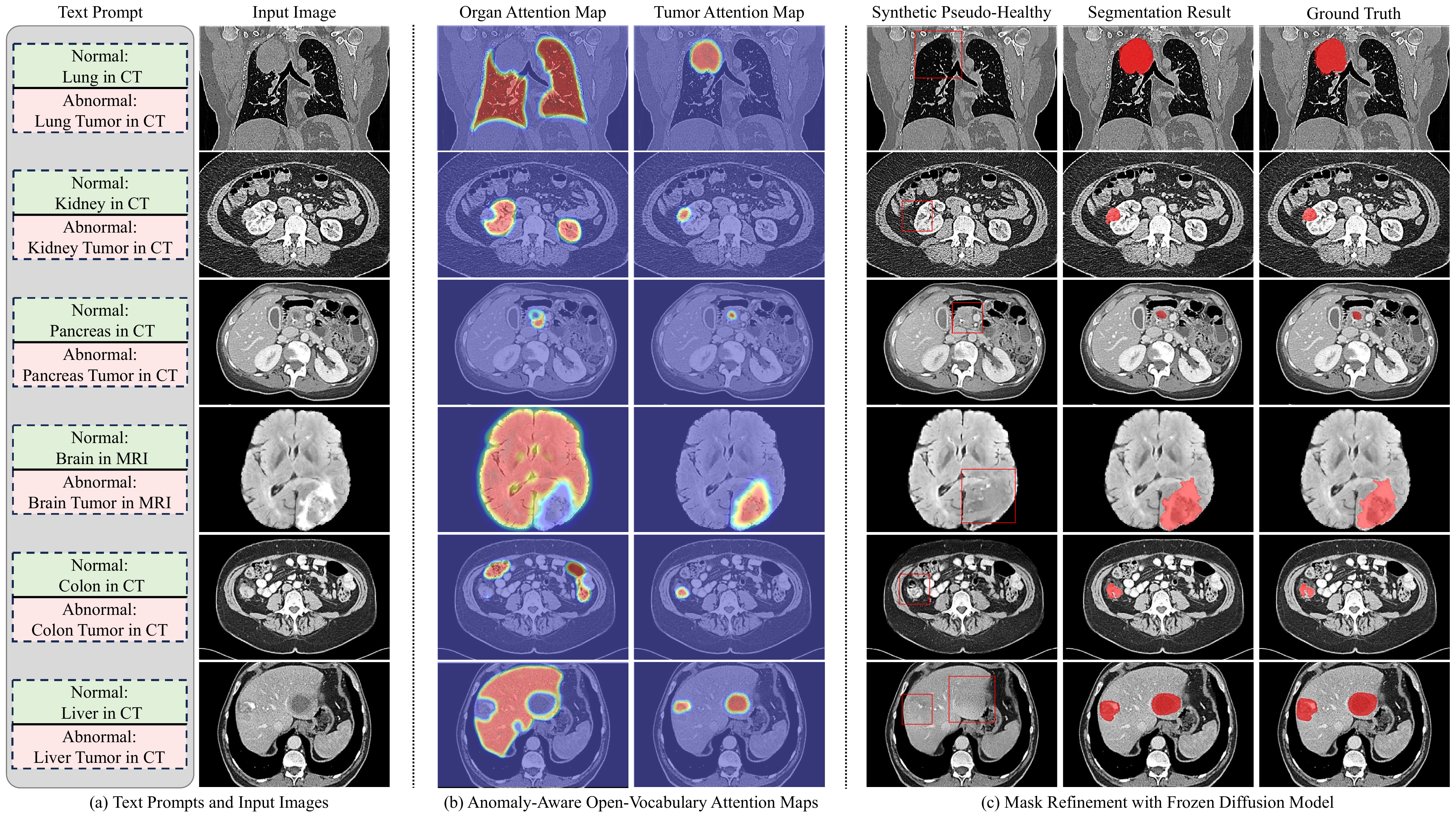}
  \caption{We propose \textbf{DiffuGTS}, a novel framework that utilizes and extends the capabilities of a frozen medical foundational diffusion model for advanced zero-shot tumor segmentation across various anatomical regions and imaging modalities. (a) \textbf{DiffuGTS} employs descriptions of both normal and abnormal categories to generate open-vocabulary text-attribution attention maps (b) for anomaly segmentation through cross-modal feature interactions. Furthermore, \textbf{DiffuGTS} leverages the frozen diffusion model to refine anomaly segmentation masks (c) by synthesizing pseudo-healthy equivalents and applying pixel-level and feature-level residual learning, significantly surpassing the performance of existing zero-shot lesion segmentation methods~\cite{jiang2024zept,huang2024adapting,jiang2024unleashing}.
}
  \vspace{-10pt}
  \label{fig:diffuGTS_intro}
\end{figure*}

With development of vision-language models (\eg, CLIP~\cite{radford2021learning}), some methods~\cite{jiang2024zept,huang2024adapting,jiang2024unleashing} have paved the way for unseen tumor segmentation through the zero-shot generalization ability of vision-language alignment between segmentation regions and text prompts. However, in medical imaging, the visual cues of tumors are often subtle and ambiguous. % and their correlation with texts is far less direct compared to image-text pairs in natural images. 
Without a large amount of high-quality image-text pairs for training, segmenting unseen tumor categories using text prompts alone is highly challenging. Furthermore, the vision-language alignment process based on contrastive learning is not necessarily optimal for pixel-level segmentation, as the training objective is not directly optimized for spatial and relational understanding. As a result, these methods are often limited in segmentation quality.

%to segment unseen tumor categories is highly challenging, resulting in suboptimal model performance.

%By creating artificial yet realistic medical images, synthetic data can augmentexisting datasets, reduce the dependency on real patient data, and provide a cost-effective alternative to manual data annotation.

Another promising direction towards GTS is tumor synthesis~\cite{hu2023label,chen2024towards,wu2024freetumor}, which enables label-free tumor segmentation by creating artificial
yet realistic medical images. %With the emergence of latent diffusion models (LDMs)\cite{rombach2022high}, an increasing number of methods\cite{chen2024towards,wu2024freetumor,guo2024maisi} are capable of creating realistic tumors that generalize across a range of organs. AI models trained on these synthetic tumors have demonstrated the ability to detect and segment real tumors from CT or MRI volumes. 
However, current tumor synthesis methods are unable to encompass all tumor types, as simulating tumors with irregular shapes or those not encountered during diffusion model training poses significant challenges~\cite{wu2024freetumor,guo2024maisi}. Consequently, achieving GTS relying solely on synthetic medical data remains an intricate problem.

Motivated by the above challenges, we take an innovative approach towards GTS. %We observe that, %regardless of the realism of synthetic data, 
Our key insight is that, despite the challenges in simulating tumors, a medical foundation diffusion model (MFDM) trained on large-scale data is capable of learning and understanding rich, diverse anatomical structures and organ-specific knowledge~\cite{guo2024maisi}. Moreover, this valuable knowledge is already embedded within its internal representations.
%medical foundation diffusion models (MFDMs), trained on large-scale real data, are capable of learning and understanding rich, diverse anatomical structures and organ-specific knowledge~\cite{guo2024maisi}. 
Therefore, instead of synthesizing data for model training, we uncover the potential of pre-trained MFDMs as highly efficient semantic feature extractors and demonstrate that their internal image representations can be repurposed, through carefully designed strategies, as effective zero-shot learners for the GTS task.

To this end, we propose \textbf{DiffuGTS}, a novel framework with strong zero-shot capabilities that leverages frozen MFDMs for generalizable tumor segmentation. We utilize the cross-modal interactions between the internal visual representations of frozen MFDMs and text prompts for anomaly detection to construct a set of novel anomaly-aware open-vocabulary attention (AOVA) maps to achieve zero-shot tumor detection. The AOVA maps repurpose and recalibrate the diverse anatomical knowledge from MFDMs for unseen anomaly segmentation, improving efficiency and enabling generalization to unseen image modalities.%by eliminating the need for training a versatile image encoder with vision-language alignment, as required in previous zero-shot tumor segmentation methods~\cite{jiang2021ala,jiang2024unleashing}.
%recalibrate the model’s focus from object semantics in natural imagery
%generalizes the use of attention maps from cross-modal interaction between the internal visual representation of medical foundation diffusion models and diagnostic text prompts.

Furthermore, we leverage the generative capabilities of frozen MFDMs to refine the anomaly segmentation masks derived from AOVA maps. We first adopt a training-free latent space inpainting strategy that transforms pathological regions into pseudo-healthy equivalents, conditioned on the anomaly segmentation masks. Then, a novel pixel-level and feature-level residual learning approach generates refined segmentation masks by identifying discrepancies between the original pathological regions and their corresponding pseudo-healthy equivalents, enabling substantial advancements in zero-shot tumor segmentation performance.
%which can preserve details of the input image that are not affected by the disease while re-painting the diseased part with realistic looking tissue. We then propose pixel-level and feature-level residual learning to capture residuals between the original diseased organs and pseudo-healthy equivalents on the fly, our model can gain a cohesive understanding of diverse anomalies, enabling remarkable zero-shot tumor segmentation performance.

Extensive experiments across multiple datasets validate the superiority of \textbf{DiffuGTS} (\cref{fig:diffuGTS_intro}), significantly surpassing previous SOTA methods under various challenging zero-shot settings. In a nutshell, our work offers: (1) an effective and efficient framework with novel designs capable of segmenting unseen tumors across diverse anatomical regions and imaging modalities in a zero-shot manner; (2) superior performance: comparisons with various methods across four datasets demonstrate that our method establishes a new state-of-the-art for generalizable tumor segmentation; (3) in-depth analysis: multiple visualizations imply the strong zero-shot capabilities of AOVA maps for anomaly detection and the efficacy of mask refinement with frozen MFDMs, while key ablation experiments demonstrate the effectiveness of our strategies.

\section{Related Work}
\label{sec:relatedwork}
\noindent \textbf{Zero-Shot 3D Medical Image Segmentation.}
Existing zero-shot 3D medical image segmentation methods fall primarily into two categories: SAM~\cite{kirillov2023segment}-based methods~\citep{wald2023sam,yamagishi2024zero,aleem2024test,shaharabany2024zero} and methods based on  vision-language alignment~\cite{jiang2024zept,jiang2024unleashing}.
SAM-based 3D medical image segmentation methods~\citep{wald2023sam,yamagishi2024zero,aleem2024test,shaharabany2024zero} demonstrate promising zero-shot performance in segmenting certain organs, particularly larger organs with clear boundaries. However, they primarily focus on organ segmentation and have \textbf{not} been evaluated or proven effective when confronted with unseen lesions that have less defined structures. Recently, some methods~\cite{jiang2024zept,jiang2024unleashing}, achieve competitive language-driven zero-shot tumor segmentation performance by matching mask proposals with text descriptions through contrastive learning. However, relying only on weak supervision from text descriptions, these methods are limited to compromised zero-shot performance. %As show in fig.4, we observe language-driven anomaly segmentation maps struggle to capture the precise contours of tumors. 
\textbf{DiffuGTS} takes a huge step further by leveraging frozen MFDMs to achieve significantly superior zero-shot tumor segmentation performance across different image modalities and various anatomical regions.
%based on language-driven anomaly segmentation maps, demonstrating its exceptional generalizability across diverse data modalities and anatomical regions.  %To the best of our knowledge, FDM-GTS is the first zero-shot method to achieve comparable, or even superior, zero-shot tumor segmentation performance compared to a strong supervised baseline nnUNet~\citep{isensee2021nnu}.
%The model's ability to segment previously unseen targets without additional training highlights its potential for cross-domain generalization in medical imaging. 

%---------------------------------------------------------------
\noindent \textbf{Diffusion Models for Medical Image Segmentation.}
Diffusion models have recently demonstrated significant potential in various medical image segmentation tasks~\cite{kim2022diffusion,wolleb2022diffusion,guo2023accelerating,rahman2023ambiguous,wu2024medsegdiff,wu2024medsegdiffv2}. The majority of diffusion-based segmentation methods ~\cite{wolleb2022diffusion,guo2023accelerating,rahman2023ambiguous,chowdary2023diffusion,chen2023berdiff, wu2024medsegdiff,wu2024medsegdiffv2} focus on enhancing the segmentation quality of specific organs or tissues under fully supervised settings. Some methods~\cite{chen2024towards,wu2024freetumor,guo2024maisi}, on the other hand, focus on generating additional medical data along with corresponding annotations to supple training data. 
%This approach helps alleviate data scarcity issues and enhances segmentation performance. 
However, tumor synthesis remains a challenging issue, primarily due to concerns about the quality of synthetic data and the limited diversity of synthesizable categories.
%but also because existing methods are unable to generate unseen tumor types that were not encountered in the training set. %To the best of our knowledge, \textbf{DiffuGTS} is the first framework tailed for zero-shot tumor segmentation with diffusion models.
In contrast, \textbf{DiffuGTS} focuses on leveraging the frozen diffusion models for zero-shot tumor segmentation.
%by modeling the tumor segmentation as a residual learning process between 
%Moreover, the diffusion model’s hierarchical structure makes it possible to govern the ambiguity at each time step, thereby eliminating the problem of low diversity of previous methods.
%---------------------------------------------------------------

\noindent \textbf{Diffusion Models for Medical Anomaly Detection.}
Diffusion models have shown substantial promise in enhancing the precision of medical anomaly detection by transforming pathological inputs into pseudo-healthy outputs and then computing the difference between the original and synthetic images~\cite{wolleb2022diffusion2, bercea2024diffusion}. Current methods~\cite{wyatt2022anoddpm, behrendt2023guided, wolleb2022diffusion2, bercea2024diffusion} mainly focus on 
enhancing the quality of generated pseudo-healthy outputs.
%improving the model's ability to preserve details of the input image that are not affected by the disease while transforming the diseased part into realistic-looking tissue. 
However, these methods are tailored for specific anatomical regions (\textit{e}.\textit{g}., brain or chest) and are restricted to handling particular tumor categories, failing to generalize across diverse diseases and image modalities—a critical aspect that our paper seeks to address.

\section{Method}
\begin{figure*}
  \centering
  \includegraphics[width=\linewidth]{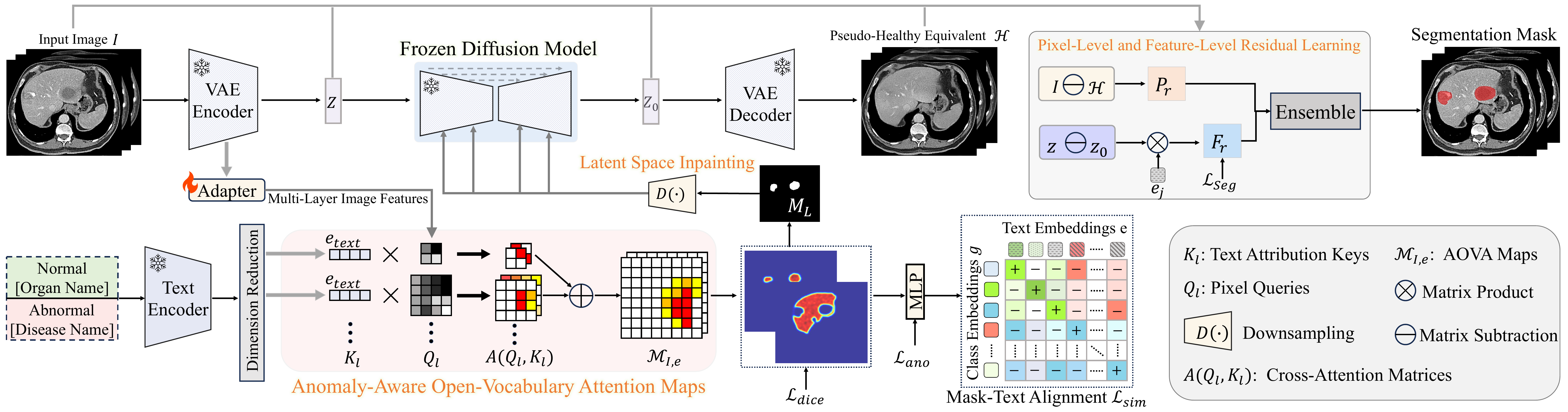}
  \caption{The architecture of \textbf{DiffuGTS} for generalizable tumor segmentation.}
  \vspace{-10pt}
  \label{fig:framework}
\end{figure*}

{\bf DiffuGTS} first explores internal representations from a frozen foundational diffusion model, MAISI~\cite{guo2024maisi}, to efficiently leverage anatomical features and create anomaly-aware open-vocabulary attention (AOVA) maps for tumor detection (Sec.\ref{subsec3.1}). Subsequently, it employs the frozen foundational diffusion model to synthesize pseudo-healthy images conditioned on the AOVA maps, allowing for the extraction of tumor segmentation masks by analyzing the pixel-level and feature-level discrepancies between the original diseased images and their pseudo-healthy counterparts (Sec.\ref{subsec3.2}). \cref{fig:framework} illustrates the pipeline of {\bf DiffuGTS}. We elaborate on the details of our design in the following. %We provide the preliminaries, discussing MAISI and the problem formulation of zero-shot lesion segmentation, as prior information required for our work in Appendix~\ref{Preliminaries}.

%Denoising diffusion probabilistic models (DPM) can also serve as a representation learner to capture semantic information and potentially be used as an image representation for downstream tasks, \eg, segmentation.
%improves semantic representation through inexpensive medical text annotations, thereby explicitly establishing semantic representation and language correspondence for diffusion models.

%We aim to leverage frozen foundational diffusion models (MFDMs)~\cite{guo2024maisi} for efficient and precise zero-shot tumor segmentation. Synthesizing training data with realistic tumors for all target disease categories is challenging even for current MFDMs. Conversely, masking out the lesion regions and allowing the MFDMs to synthesize healthy organ tissues is relatively more feasible. Therefore, segmenting tumors can be achieved by analyzing the differences between the pathological input image and its synthetic healthy equivalent. 
%Tumor synthesis enables the creation of artificial tumors in medical images, facilitating the training of AI models for tumor detection and segmentation
%Furthermore, since the MFDMs have been trained on a large amount of data encompassing different modalities and organs, they have acquired rich knowledge of various anatomical structures, enabling us to exploit their internal representations for generalizable tumor segmentation across different diseases and anatomical regions in a zero-shot manner.

\subsection{Formulation of AOVA Maps}
\label{subsec3.1}
To facilitate generalizable tumor localization across diverse anatomical regions, we introduce anomaly-aware open-vocabulary attention maps, which allow us to control attention heatmaps for constructing anomaly segmentation masks using text prompts (see~\cref{fig:diffuGTS_intro}). This approach establishes a direct correlation between the spatial anatomical layout and the semantic content of diagnostic text descriptions, eliminating the constraints imposed by a predefined category list during training. As a result, it enables zero-shot generalization to previously unseen tumor categories.

%establishing a direct correlation between the spatial layout and the semantic content of the text.
%thereby explicitly establishing a correspondence between semantic representations and language.
%We create the anomaly-aware open-vocabulary attention maps for the zero-shot tumor detection by exploring the latent alignment of semantic features with diagnostic descriptions. 

%\subsection{Text-Prompted Zero-Shot Anomaly Detection}
%increase the data usability by extracting knowledge from other readily available sources of medical information, such as text diagnostic information, to complement medical images.

\noindent \textbf{Visual Feature Extraction.}
%These embeddings capture semantic information from both the original descriptions and the LLM-generated knowledge, enabling the model to better differentiate between organ and lesion categories in the downstream segmentation tasks.
In contrast to existing zero-shot tumor segmentation methods~\cite{jiang2024zept,jiang2024unleashing}, which typically require the training of versatile vision encoders, we directly and efficiently utilize feature representations from a frozen foundational diffusion model.
%mines the latent alignment of semantic features in diffusion models with diagnostic descriptions by only training the cross-attention mechanism and pixel classifier, making it possible to enhance semantic representation with inexpensive text.
Specifically, we exploit the internal image features from MAISI's VAE encoder~\cite{guo2024maisi}. For an input 3D volume $\mathcal{I}$, the MAISI VAE encoder transforms $\mathcal{I}$ into multi-scale image features $F_l \in \R^{H_l \times W_l \times D_l \times C_l}, l \in \{1,2,3\}$. Here, $l$ denotes the three internal stages, while $H_l$, $W_l$, $D_l$, and $C_l$ represent the height, width, depth, and channel dimensions of $F_l$, respectively. 

Since we fix the parameters of the VAE encoder $V_E$ which is not originally optimized for segmentation tasks, there is a potential gap between the generative and discriminative representation space. Therefore, we perform visual feature adaptation to tailor representations for segmentation tasks, while maintaining the VAE encoder's rich anatomical knowledge. At each level $l$, a learnable feature adapter, encompassing two layers of linear transformations and a residual connection, projects the image features $F_l$ for adaptation, represented as: $\gF_l = \alpha T_l(F_l) + (1-\alpha) F_l$. Here, $T_l(\cdot)$ denotes the learnable parameters of the linear transformations. A constant value $\alpha$ serves as the residual ratio to adjust the degree of preserving the original knowledge for improved zero-shot performance. By default, we set $\alpha = 0.1$.

\noindent \textbf{Textual Prompt Composition.} To explicitly provide hints and prompt the model for effective anomaly detection, we adopt text prompts for the construction of cross-modal correlations between vision and language features. Specifically, we leverage descriptions of both normal and abnormal categories. For normal organs, the predefined template is: ``A nomal CT scan/MRI of \{organ name\}.'' For tumors, the template is: ``An abnormal CT scan/MRI of \{disease name\}.'' %Inspired by recent works~\cite{zhang2023knowledge,wu2023medklip,jiang2024zept}, which demonstrate that incorporating common knowledge enhances medical vision-language models, we also utilize the Large Language Model (LLM) GPT-4~\citep{achiam2023gpt} to provide versatile and generic descriptions regarding the location, shape, and specific features for the normal and abnormal categories, serving as a supplement of medical domain knowledge to the text prompts. %We concatenate the text prompts with their corresponding knowledge and 
%We concatenate the normal and abnormal text prompts and then extract text features $\mathcal{T} \in \R^{2 \times d}$ using a pre-trained frozen text encoder, where $d$ denotes the feature dimension.
We extract text embeddings $e \in \R^{N \times d}$ from normal and abnormal prompts using a pre-trained frozen text encoder~\cite{yan2022clinical}, where $N$ and $d$ denotes the number of training categories and feature dimension.

%\subsection{Open-Vocabulary Segmentation Masks}
%\label{subsec3.2}
\noindent \textbf{Text-Driven Region-Level Anomaly Detection.}
For the construction of AOVA maps, we use the features of text prompts to control the attention heatmaps derived from the cross-attention matrices. At each image feature level $l \in \{1,2,3\}$, the text features first undergo dimension reduction through a Multi-Layer Perception (MLP), adapting them to be compatible with the image features in terms of dimension size and thus enhances computational efficiency. Then, the text embeddings $e$ are projected using the key projections $\mathcal{W}^l_K$ to generate attribution keys $K_l = \mathcal{W}^l_K(e) \in \R^{N \times C_l}$. Similarly, we project the image features $\gF_l$ using the query projections $\mathcal{W}^l_Q$ to generate pixel queries $Q_l = \mathcal{W}^l_Q(\gF_l) \in \R^{H_l \times W_l \times D_l \times C_l}$. The attribution keys $K_l$ are combined with the pixel queries $Q_l$, creating the cross-attention matrices $A(Q_l, K_l) = \operatorname{Softmax}\left(\frac{Q_l K_l^T}{\sqrt{C_l}}\right) \in \R^{H_l \times W_l \times D_l \times N}$.
%\begin{equation}
%\centering
%   A(Q_l, K_l) = \operatorname{Softmax}\left(\frac{Q_l K_l^T}{\sqrt{d}}\right) \in \R^{H_l \times W_l \times D_l \times N}.
%  \label{eq:2}
%\end{equation}
These cross-attention matrices weigh the influence of text features for both normal and abnormal objects on the image's pixels, establishing a direct correlation between the anatomical spatial layout and the semantic content of the descriptions for normal and abnormal categories.

We aggregate cross-attention matrices across $l$ feature levels to generate the AOVA maps as:
%To generate the final text-prompted attention maps relevant to the $N$ classes, we aggregate the matrices across different feature levels as:
\begin{equation}
\centering
   M_{\mathcal{I}, e} = \sum_{l=1}^{3} \mathbf{RI} \big(A(Q_l, K_l)\big) \in \R^{H \times W \times D \times N}.
  \label{eq:3}
\end{equation}
Here, $\mathbf{RI}(\cdot)$ denotes the reshape operation using bilinear interpolation for resizing $A(Q_l, K_l)$ of varying resolutions to the original input image resolution. 

%Considering that tumor segmentation poses significantly more challenges than organ segmentation, To further improve the We also utilize images from healthy individuals to construct a memory bank $\gB$ that stores multi-level features of healthy organs, facilitating feature comparison for anomaly detection. The anomaly segmentation maps $A_u(\gF_l) \in \R^{H \times W \times D \times N_u}$ for the $N_u$ tumor classes are derived from the minimum distance between the image features and the memory bank features at each level, through a nearest neighbor search process:
%\begin{equation}
%\centering
%   A_u(\gF_l) = \frac{3}{1} \sum_{l=1}^{3} \operatorname{resize} \big(\min_{f \in \gB} Dist(\gF_l,f)\big).
%  \label{eq:4}
%\end{equation}
%Here, $Dist(\cdot,\cdot)$ represents the cosine distance, calculated as $1-\operatorname{cosine}(\cdot,\cdot)$. The open-vocabulary segmentation probability maps are obtained by combing the text-prompted attention maps $M_{\mathcal{I}, t}$ and the anomaly segmentation maps $A_u(\gF_l)$: 
%\begin{equation}
%\centering
%  \gM_{\mathcal{I}} = 
%  \begin{cases}
%       M_{\mathcal{I}, t}[:N_u] + A_u(\gF_l) \text{ } \in \R^{H \times W \times D \times N_u} \\%& \mbox{for organs} \\
%       M_{\mathcal{I}, t}[N_u:N]  \text{  } \in \R^{H \times W \times D \times N_o} .%& \mbox{otherwise}
%  \end{cases}
% \label{eq:5}
%\end{equation}
%Here, $N_u$ and $N_o$ denote the number of tumor and organ classes, respectively ($N = N_u + N_o$).

%\textbf{Mask Classification.}
We optimize the AOVA maps $M_{\mathcal{I}, e}$ for anomaly detection through 
three training objectives.
%establishing alignment between the attention maps and text representations within the multi-modal feature space.
First, we feed each AOVA map into a MLP layer to obtain the class embedding $g_i \in \R^{d}, i \in [1,N]$. We use the $g_i$ to predict an anomaly score $ascore = \operatorname{Sigmoid}(MaxPool(g)): \mathbb{R}^d \to [0,1]$ so that a binary classification can be performed using binary cross-entropy loss: $\gL_{ano} = \operatorname{BCE}(th(ascore),\mathcal{C})$. Here, $th(\cdot)$ denotes a threshold set to 0.5, and $\mathcal{C} \in \{-,+\}$ represents the image-level anomaly annotation, where '$+$' indicates an anomalous sample and '$-$' denotes a normal one. 

Then, for AOVA maps classified as abnormal, we align their corresponding class embeddings $g_i$ with the text embeddings of abnormal categories. Conversely, for maps classified as normal, we align their class embeddings $g_i$ with the text embeddings of normal categories.
The alignment is achieved by a CLIP-style contrastive learning~\cite{radford2021learning} approach. %Let the text embeddings for each category be denoted as $e_i \in \mathbb{R}^d$. 
The similarity score between each class embedding and text embedding is computed by a dot product normalized by a temperature parameter $\tau$: $s(g_i, e_j) = \frac{g_i \cdot e_j}{\tau}$. Then the similarity score is refined through a contrastive loss function defined as:
\begin{equation}
\centering
%\resizebox{0.9\linewidth}{!}{$
  \gL_{sim} = -\frac{1}{N} \sum_{i=1}^{N}   \log \frac{\exp \big(s(g_i, e_j)\big)}{\sum_{j=1}^{N} \exp \big(s(g_i, e_j)\big)}.
%$}
  \label{eq:6}
  %\vspace{-3pt}
\end{equation}
This contrastive learning aligns the AOVA maps with text embeddings, enabling open-set anomaly detection that generalizes to unseen categories.

Moreover, we convert $N$ AOVA maps into $N$ binary segmentation masks for normal and abnormal categories through a $\operatorname{Sigmoid}$ operation, and supervise these masks with partially labeled segmentation annotations $\mathcal{S}$ using a $\operatorname{Dice}$ loss~\cite{milletari2016v}: $\gL_{dice} = \operatorname{Dice}(\operatorname{Sigmoid}(M_{\mathcal{I}, e}), \mathcal{S})$. %Here, $\mathcal{S} \in \{-,+\}^{H\times W\times D\times N}$.
During training, text descriptions for all categories are provided, and healthy cases for every organ are utilized, enabling the network to differentiate normal organs from tumors and thereby achieve zero-shot generalization for lesions.
The overall loss function for AOVA map optimization is: $\gL_{AOVA} = \lambda_1 \gL_{ano} + \lambda_2 \gL_{sim} + \lambda_3 \gL_{dice}$. 
%where $\lambda_1$, $\lambda_2$, and $\lambda_3$ are weighting factors balancing the contribution of each loss component to the overall training objective. 
We set $ \lambda_1 = \lambda_2 = \lambda_3 = 1.0$ as default.

\subsection{Mask Refinement with Frozen Diffusion Model}
\label{subsec3.2}
The AOVA maps can accurately localize tumors using open-vocabulary text prompts, as shown in~\cref{fig:diffuGTS_intro}. However, during zero-shot inference, the masks generated by such text-driven anomaly detection are inherently limited in quality due to the absence of fine-grained, pixel-level supervision for unseen tumor categories.
%the mask generated by text-driven anomaly detection has inherent limitations in terms of mask quality. 
To further enhance zero-shot segmentation performance, we propose utilizing a frozen latent diffusion model from MAISI~\cite{guo2024maisi} to refine the lesion masks. This is achieved by synthesizing a pseudo-healthy equivalent of the target diseased organ through an inpainting task. The tumor region can then be obtained by computing the discrepancies between the input image and the synthetic image. The tumor mask refined through this process is significantly more precise than the anomaly segmentation map.

\begin{figure}
  \centering
%  %\fbox{\rule{0pt}{2in} \rule{0.9\linewidth}{0pt}}
  \includegraphics[width=\linewidth]{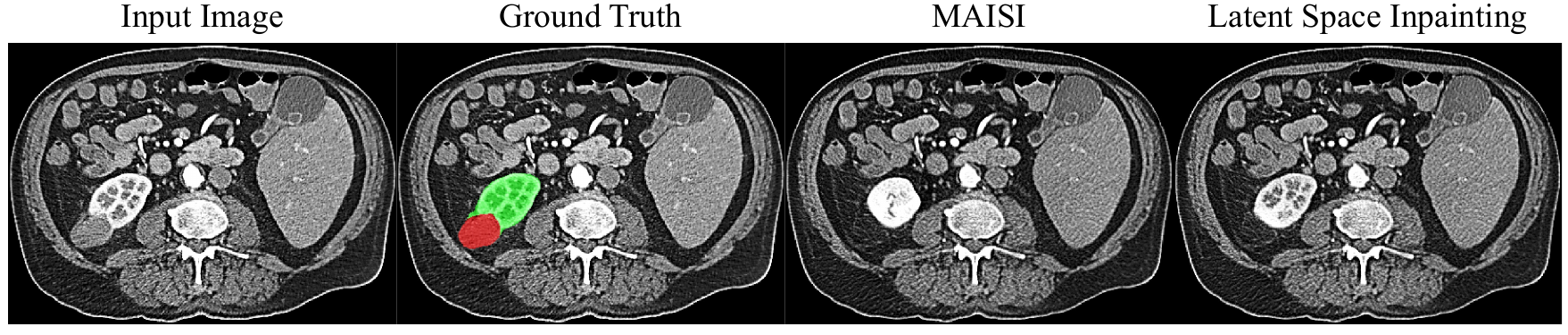}
  \caption{Synthesizing pseudo-healthy images directly using MAISI or utilizing the training-free latent space inpainting.
  }
  \vspace{-15pt}
  \label{fig:Latent}
\end{figure}

\noindent \textbf{Training-Free Latent Space Inpainting.} 
MAISI~\cite{guo2024maisi} integrates a ControlNet~\cite{zhang2023adding} to synthesize a healthy organ conditioning on the given organ mask. However, as shown in~\cref{fig:Latent}, it does not ensure that the organ regions unaffected by the disease are preserved, which significantly hinders the performance of tumor segmentation based on differences between images. 
A key aspect of anomaly segmentation is to ensure fidelity to the original scan in areas unaffected by pathology~\cite{wolleb2022diffusion2,bercea2024diffusion}. To achieve this, we reformulate the original conditional generation process of MAISI as latent space inpainting, leveraging the anomaly segmentation mask as conditions to guide the organ synthesis.

%inpainting~\cite{chen2024towards}. Since our goal is to transform the tumor region into normal healthy tissue, we need to use the mask of the entire diseased organ as a condition for healthy organ synthesis, rather than only using the mask of the abnormal region, which is suitable for tumor synthesis.
%We leverage the capability of MAISI~\cite{guo2024maisi} to synthesize healthy organs given a corresponding organ mask.
Let the latent representation of the input 3D volume $\mathcal{I}$, generated by the VAE encoder $V_E$, be denoted as $z = V_E(\mathcal{I}) \in \R^{h \times w \times d \times c}$. To compensate for the issue that anomaly segmentation maps of unseen tumors cannot always precisely capture tumor boundaries, we first enlarge the tumor region in the anomaly segmentation mask using a coefficient $\beta$ to ensure that the entire tumor region can be covered by the mask.
%corresponding organ mask to obtain a enlarged mask $M_{\text{L}} \in \mathbb{R}^{H \times W \times D}$ that covers the entire diseased organ region. 
Then, we extend the training-free strategy proposed in~\cite{lugmayr2022repaint}, which enables conditioning the inpainting task on the known region, thereby eliminating the need for fine-tuning the MAISI~\cite{guo2024maisi} model.
Specifically, we obtain an intermediate latent code by regenerating the potential tumor region from the model's output while sampling other normal regions from the input. The reverse step at timestep $t$ is formulated as follows:
\begin{align}
\centering
   z_{t-1}^{\text{other}} &\sim \mathcal{N}\left(\sqrt{\overline{\alpha}_t}z, (1-\overline{\alpha}_t)\mathbf{I}\right),\\ 
   z_{t-1}^{\text{tumor}} &\sim \mathcal{N}\left(\mu_\theta(z_t,t), \Sigma_\theta(z_t,t)\right), \\ 
   z_{t-1} &= (1 - D(M_L))\otimes z_{t-1}^{\text{other}} + D(M_L)\otimes z_{t-1}^{\text{tumor}}.
\end{align}
Here, $\odot$ represents element-wise multiplication, $D(\cdot)$ is the downsampling operation, $\overline{\alpha}=\Pi_{s=1}^t (1-\beta_s)$ and $\beta_s$ denotes the variance of Gaussian noise at timestep $s$ according to a variance schedule predefined in MAISI\cite{guo2024maisi}. Unlike the standard RePaint~\cite{lugmayr2022repaint} method, which operates in pixel space, we adapt the enlarged mask $M_{\text{L}}$ to match the latent space dimensions by downsampling it using nearest-neighbor interpolation, as in LatentPaint  \cite{corneanu2024latentpaint}. After obtaining the final step output latent embeddings $z_0 \in \R^{h \times w \times d \times c}$, we then use the frozen VAE decoder $V_D$ in MAISI to generate the pseudo-healthy image $\mathcal{H} = V_D(z_0) \in \R^{H \times W \times D}$.
In~\cref{fig:Latent}, we show that the pseudo-healthy images generated using our extended, training-free latent space inpainting strategy preserve the healthy regions of the organ to a much greater extent, significantly outperforming the results obtained by directly applying the MAISI model for organ synthesis. We provide the illustration of our one-step reverse process in the supplementary material. 

\noindent \textbf{Pixel-Level and Feature-Level Residual Learning.} %Given the original input 3D volume $\mathcal{I}$ and its corresponding synthesized healthy variant $\mathcal{S}$, 
We first obtain the pixel-level residual map $P_r$ by performing element-wise subtraction between the normalized input volume $\mathcal{I}$ and its pseudo-healthy variant $\mathcal{H}$ : $P_r = \mathcal{I} \ominus \mathcal{H} \in \mathbb{R}^{H \times W \times D}$. %In $P_r$, higher values indicate a higher probability that the corresponding pixel belongs to the tumor category.  
This pixel-level residual map, commonly used in previous medical anomaly detection methods~\cite{siddiquee2019learning,wolleb2022diffusion2,kumar2023self,bercea2024diffusion}, offers computational efficiency and intuitively highlights pixel-level discrepancies between images.
%can be converted into an anomaly segmentation mask by applying a threshold, 
However, its accuracy is highly contingent upon the quality of the synthesized images; any inconsistencies in generation can lead to pixel variations that compromise segmentation performance. %This limitation is particularly evident when the contrast between the lesion and surrounding tissue is low, as residual learning in pixel space may fail to detect subtle lesions. 
%Moreover, determining the optimal threshold is challenging, as it varies across different tumor categories.
Thus, we propose a novel pixel-level and feature-level residual learning to obtain the final tumor segmentation masks. This method combines $P_r$ with a feature-level residual map that highlights the differences between $\mathcal{I}$ and $\mathcal{H}$ in latent space to effectively discriminate tumors.
%We additionally leverage feature-level differences between $\mathcal{I}$ and $\mathcal{H}$ to discriminate tumors in feature space. %This method enables the construction of a feature-level residual map. %and enhances the model’s anomaly-discriminative capacity. 

Specifically, we obtain the feature-level residual map by performing element-wise subtraction as follows: $f_r = z \ominus z_0$. Compare with the pixel-level residual map, $f_r$ incorporates a deeper semantic understanding, capturing complex structural and pattern changes rather than merely pixel-level brightness or density variations. Then we align $f_r$ with its corresponding text embeddings of prompts $e_j$ to obtain feature-level anomaly segmentation maps $F_r = \Psi \big(f_r \cdot \psi(e_j) \big) \in \R^{H \times W \times D}$. Here, $\Psi$ and $\psi$ represent the upsampling and linear projection operations, respectively. Finally, a Dice loss~\cite{milletari2016v} is adopted to supervise $F_r$ with segmentation annotations $\mathcal{S}$: $\gL_{Seg} = \operatorname{Dice}(\operatorname{Sigmoid}(F_r),\mathcal{S})$. During inference, we combine the pixel-level and feature-level residual maps to get the overall anomaly segmentation maps $R_{\text{pf}} = \beta_1 P_r + \beta_2 F_r$, where $\beta_1$ and $\beta_2$ are weighting factors set to 0.5 by default. We convert $R_{\text{pf}}$ into a binary segmentation mask by applying Otsu thresholding~\cite{otsu1975threshold}.

\begin{table*}[htbp]
  \centering
  %\begin{center}
  \resizebox{\linewidth}{!}
  {
  \begin{tabular}{@{}c|l|cc|cc|cc|cc|cc|cc@{}}
    \toprule
    \multicolumn{1}{c|}{\multirow{3}{*}{Method Type}} & \multicolumn{1}{l|}{\multirow{3}{*}{Method}} & \multicolumn{10}{c|}{MSD Dataset} &\multicolumn{2}{c}{KiTS23 Dataset}\\
    \cline{3-14}
     & & \multicolumn{2}{c|}{Pancreas Tumor} & \multicolumn{2}{c|}{Lung Tumor} & \multicolumn{2}{c|}{Liver Tumor} & \multicolumn{2}{c|}{Colon Tumor} & \multicolumn{2}{c|}{Hepatic Vessel Tumor} & \multicolumn{2}{c}{Kidney Tumor}\\
    \cline{3-4}\cline{5-6}\cline{7-8} \cline{9-10} \cline{11-12}  \cline{13-14}   
    & & DSC$\uparrow$ & NSD$\uparrow$ & DSC$\uparrow$ & NSD$\uparrow$ & DSC$\uparrow$ & NSD$\uparrow$
    & DSC$\uparrow$ & NSD$\uparrow$ & DSC$\uparrow$ & NSD$\uparrow$ & DSC$\uparrow$ & NSD$\uparrow$\\
    \hline
    %MaxLogit~\cite{hendrycks2022scaling}          & 80.95 & 80.95 & 80.95 & 80.95 & 80.95 & 80.95  & 80.95    & 85.23     & 81.04   & 85.60   & 81.33  &  86.05 & 80.95 & 80.95 & 80.95 \\
    \multirow{3}{*}{SAM-based Methods} & SAM 2~\cite{yamagishi2024zero}     & 26.65 & 38.77         & 15.39 & 15.96            & 42.75 & 50.22        & 13.08 & 21.40               & 38.92 & 49.08  &  36.48 & 42.59 \\
    & SaLIP~\cite{aleem2024test}     & 31.28 & 44.33            & 20.05 & 20.77          & 48.39 & 56.90         & 19.33 & 27.02               & 44.18 & 55.84   & 39.11 & 45.24 \\
    & H-SAM~\cite{cheng2024unleashing}  & 35.19 & 50.02       & 25.36 & 26.11          & 53.03 & 60.44         & 23.02 & 30.67          & 51.85 & 61.24   & 45.57  & 52.16 \\
    \hline
    \multirow{2}{*}{\makecell[c]{3D Zero-Shot Lesion \\ Segmentation Methods}} & 
     ZePT~\cite{jiang2024zept}         & 39.40 & 54.76           & 30.02 & 31.23             & 59.16 & 68.72         & 33.85 & 42.31              & 55.83 & 65.72  & 48.75 & 54.91   \\
    & Malenia~\cite{jiang2024unleashing}     & 40.26 & 55.82   & 32.75 & 33.92              & 59.83 & 70.08         & 34.72 & 42.59       & 59.71 & 69.98  & 55.37 & 61.16 \\
    \hline
    \multirow{4}{*}{\makecell[c]{Medical Anomaly \\Detection Methods}} & DDPM-MAD~\cite{wolleb2022diffusion2}     & 28.52 & 40.18           & 25.70 & 26.81             & 43.54 & 51.03         & 13.97 & 21.84              & 39.57 & 49.62   & 36.55  & 42.72\\
    & MVFA~\cite{huang2024adapting} & 32.77 & 45.63           & 24.49 & 25.07             & 49.25 & 57.72         & 20.44 & 27.87              & 45.28 & 56.14     & 40.81 & 46.29\\
    & THOR~\cite{bercea2024diffusion}     & 29.45 & 41.06           & 27.73 & 28.98           & 47.68  & 56.40         & 18.36 & 26.51              & 41.33 & 52.05    &38.39  & 44.64\\
    \cline{2-14}
     &\cellcolor{blue!8} {\bf DiffuGTS}     & \cellcolor{blue!8} \textbf{43.61} & \cellcolor{blue!8} \textbf{58.48} 
     & \cellcolor{blue!8} \textbf{42.94} & \cellcolor{blue!8} \textbf{44.01}
       & \cellcolor{blue!8} \textbf{63.23}  & \cellcolor{blue!8} \textbf{73.58} 
     & \cellcolor{blue!8} \textbf{38.72} & \cellcolor{blue!8} \textbf{45.60}  
      &\cellcolor{blue!8} \textbf{62.76}    & \cellcolor{blue!8} \textbf{72.35} 
      &\cellcolor{blue!8}  \textbf{59.80} &\cellcolor{blue!8} \textbf{65.99} \\
    \bottomrule
  \end{tabular}
  }
  %\end{center}
    \caption{Zero-shot tumor segmentation performance (\%) on MSD~\cite{antonelli2022medical} and KiTS23~\cite{heller2023kits21} with the leave-one-out setting. All the competing methods are implemented using the official code. The best result is in light blue.}
    \vspace{-10pt}
  \label{tab:MSD}
\end{table*}
\section{Experiments}
\label{results}

\noindent \textbf{Dataset Construction.}
%\noindent \textbf{Training Dataset of Base Categories.}
We consider both public and private datasets encompassing 6 organs and 7 tumor categories, sourced from multiple centers. These datasets include MSD~\cite{antonelli2022medical}, KiTS23~\cite{heller2023kits21}, BraTS23~\cite{adewole2023brain}, and an in-house MRI liver tumor segmentation dataset. We also utilized data from $404$ patients who showed no signs of pathology from the TotalSegmentator~\cite{wasserthal2023totalsegmentator} dataset as normal samples for anomaly detection training. The total number of CT and MRI scans used for training and testing in our study is $3,933$. We adopted various zero-shot testing settings for evaluation. Detailed descriptions of these datasets and the pre-processing are provided in the supplementary material.

\noindent \textbf{Evaluation Metrics.} The Dice Similarity Coefficient (DSC) and Normalized Surface Distance (NSD) are utilized to evaluate tumor segmentation performance. For all evaluation metrics, $95\%$ CIs were calculated, and a $p$-value cutoff of less than $0.05$ was used to define statistical significance.
%We adopt standard segmentation metrics, including the Dice Similarity Coefficient (DSC) and Normalized Surface Distance (NSD). 
%Additionally, the computational efficiency comparison is detailed in the Supplementary Material.
%and the false-positive rate at $95\%$ recall (FPR95).
%\noindent \textbf{Testing Dataset of unseen Categories.}
%for training：
%totalsegmentor：in its training set, A total of 404 patients showed no signs of pathology, whereas 645 showed different types of pathology (tumor, vascular, trauma, inflammation, bleeding, other).can be put in supplementary materials. (using healthy samples to avoid nosiy label problem).
\begin{figure*}
  \centering
  \includegraphics[width=\linewidth]{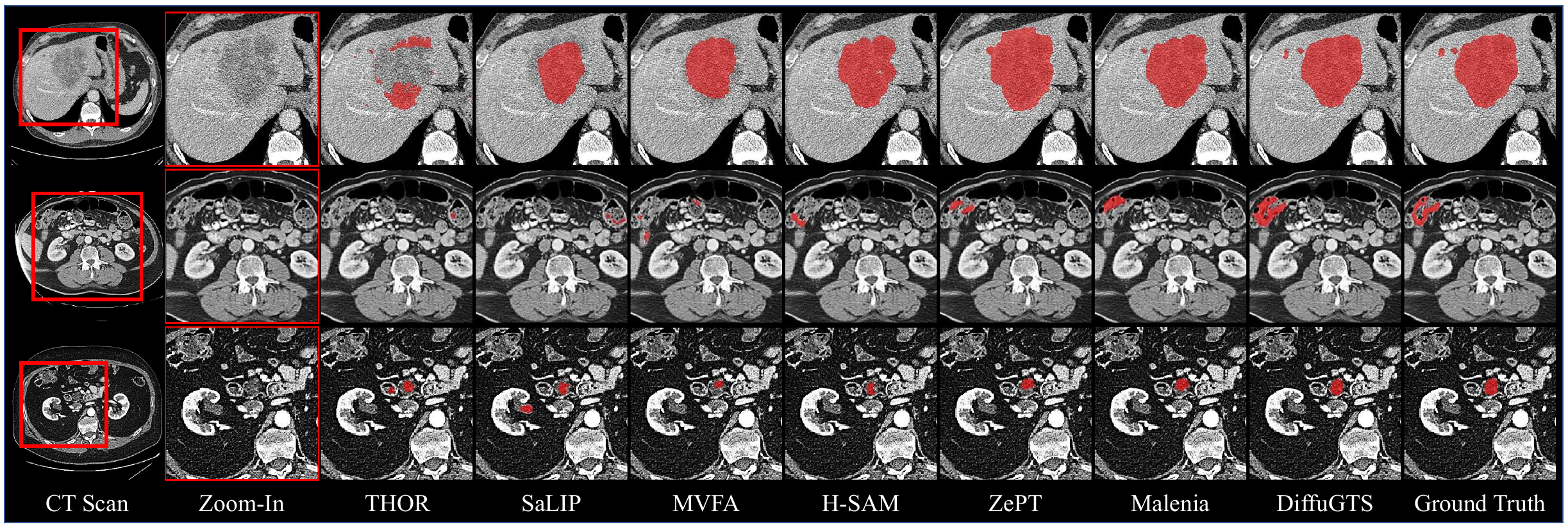}
  \caption{Qualitative visualizations of zero-shot segmentation results on MSD~\cite{antonelli2022medical}. The results presented from rows one to three correspond, in order, to liver tumors, colon tumors, and pancreatic tumors. We present the visualizations on other datasets in the supplemental material.}
  \vspace{-10pt}
  \label{fig:QR}
\end{figure*}

\noindent \textbf{Implementation Details.}
We utilize MAISI~\cite{guo2024maisi} as the frozen foundational diffusion model, which is capable of conditional generation from segmentation masks of 127 anatomical structures. We leverage the frozen 3D VAE encoder in MAISI as the feature extraction backbone.
The overall loss for training {\bf DiffuGTS} is $\gL = \gL_{AOVA} + \gL_{Seg}$.
We employ AdamW optimizer~\cite{loshchilov2017decoupled} with a warm-up cosine scheduler of $50$ epochs. The batch size is set to $2$ per GPU with a patch size of $128 \times 128 \times 128$. The training process uses an initial learning rate of $1e^{-4}$, a momentum of $0.9$, and a weight decay of $1e^{-5}$, running on $4$ NVIDIA A100 GPUs with DDP for 1000 epochs. 

\noindent \textbf{Competing Methods and Baselines.}  
In this study, we consider various state-of-the-art methods which could segment unseen tumors in zero-shot settings as competing methods, including: (i) SAM~\cite{kirillov2023segment,ravi2024sam}-based methods (Adapted SAM 2~\cite{yamagishi2024zero}, SaLIP~\cite{aleem2024test}, and H-SAM~\cite{cheng2024unleashing}), which need manual or automatic-generated prompts during testing.
(ii) 3D Zero-shot tumor segmentation methods (ZePT~\cite{jiang2024zept} and Malenia~\cite{jiang2024unleashing}), which rely on vision-language alignment. (iii) Medical anomaly detection methods (DDPM-MAD~\cite{wolleb2022diffusion2}, MVFA~\cite{huang2024adapting}, and THOR~\cite{bercea2024diffusion}).

\begin{table}
  \centering
  %\begin{center}
  \resizebox{0.98\linewidth}{!}
  {
  \begin{tabular}{@{}l|cc|cc|cc@{}}
    \toprule
    \multicolumn{1}{l|}{\multirow{3}{*}{Method}} & \multicolumn{2}{c|}{In-house} &\multicolumn{2}{c|}{MSD-Brain} &\multicolumn{2}{c}{BraTS23} \\
    \cline{2-7}
     & \multicolumn{2}{c|}{Liver Tumor} & \multicolumn{2}{c|}{Brain Tumor} & \multicolumn{2}{c}{Brain Tumor} \\
    \cline{2-3}\cline{4-5}\cline{6-7}  
    & DSC$\uparrow$ & NSD$\uparrow$ & DSC$\uparrow$ & NSD$\uparrow$ & DSC$\uparrow$ & NSD$\uparrow$ \\
    \hline
    SAM 2~\cite{yamagishi2024zero}     & 15.43 & 19.50         & 10.04 & 12.52            & 8.65 & 11.39        \\
    SaLIP~\cite{aleem2024test}     & 21.65 & 25.19            & 17.33 & 19.65          & 15.24 & 18.48        \\
    H-SAM~\cite{cheng2024unleashing}  & 26.24 & 29.37            & 19.50 & 21.97          & 17.69 & 21.45       \\
    \hline
    ZePT~\cite{jiang2024zept}     & 28.03 & 33.15           & 19.54 & 22.02             & 17.87 & 21.69      \\
    Malenia~\cite{jiang2024unleashing}     & 29.03 & 33.87           & 19.83 & 22.58             & 17.94 & 21.95     \\
    \hline
    DDPM-MAD~\cite{wolleb2022diffusion2}     & 15.78 & 19.59           & 16.11 & 18.74             & 15.08 & 17.51       \\
    MVFA~\cite{huang2024adapting}        & 22.56 & 26.74           & 19.42 & 21.89            & 17.47 & 21.19      \\
    THOR~\cite{bercea2024diffusion}     & 16.47 & 20.11           & 18.35 & 20.06             & 17.04 & 20.25   \\
    \hline
    \cellcolor{blue!8} {\bf DiffuGTS}     & \cellcolor{blue!8} \textbf{50.31} & \cellcolor{blue!8} \textbf{54.28} 
     & \cellcolor{blue!8} \textbf{47.51} & \cellcolor{blue!8} \textbf{49.75}  
      & \cellcolor{blue!8} \textbf{44.70} & \cellcolor{blue!8} \textbf{46.21}  \\
    \bottomrule
  \end{tabular}
  }
  %\end{center}
    \caption{Zero-shot tumor segmentation performance (\%) of unseen modalities on the in-house MRI liver tumor dataset, MSD-Brain~\cite{antonelli2022medical}, and BraTS23~\cite{adewole2023brain}. The best result is in light blue.}
    \vspace{-10pt}
  \label{tab:modality}
\end{table}

\subsection{Main Results}
We compare \textbf{DiffuGTS} with a series of representative SOTA methods under various zero-shot settings to assess their generalizability to unseen tumors across various anatomical regions and imaging modalities. %previously unencountered datasets

\noindent \textbf{Generalization to Unseen Tumors.}
We conducted experiments for zero-shot generalization to unseen tumors across different anatomical regions under a leave-one-out setting. In this configuration, we considered the KiTS23~\cite{heller2023kits21} along with five tumor segmentation datasets from MSD~\cite{antonelli2022medical}, including liver, colon, pancreas, lung, and hepatic vessel tumors. Each dataset was designated in turn as the unseen category (left out), while the remaining datasets were used for training. This approach allowed us to gauge the model's performance when confronted with various unseen tumor categories, thereby assessing its capacity for generalization. 

The results are shown in~\cref{tab:MSD}.
Compared with the competing SAM-based methods, \textbf{DiffuGTS} demonstrates a notable performance enhancement, achieving at least a $12.84\%$ improvement in DSC and a $13.22\%$ increase in NSD.
Due to the lack of specific knowledge about unseen tumors and the fragility of prompts, most SAM-based methods fall short in zero-shot tumor segmentation. Additionally, in real-world scenarios, obtaining precise prompts—such as points or bounding boxes derived from ground truth—is extremely challenging, further limiting the applicability of SAM-based methods. \textbf{DiffuGTS} also outperforms zero-shot lesion segmentation methods ZePT~\cite{jiang2024zept} and Malenia~\cite{jiang2024unleashing} by a large margin of at least $4.74\%$ in DSC. Although ZePT and Malenia align mask regions with textual descriptions and knowledge, relying solely on weak supervision from vision-language alignment signals is insufficient to accurately capture the boundaries of unseen tumors. In contrast, \textbf{DiffuGTS} builds upon similar vision-language alignment, while further leveraging the capabilities of the diffusion model to refine language-driven segmentation results, significantly improving performance.

Moreover, \textbf{DiffuGTS} maintains a substantial lead, with a $15.80\%$ improvement in DSC compared to medical anomaly detection methods DDPM-MAD~\cite{wolleb2022diffusion2}, MVFA~\cite{huang2024adapting}, and THOR\cite{bercea2024diffusion}. These results indicate that the proposed AOVA maps, combined with the utilization of the foundational diffusion model, effectively and significantly improve the model's accuracy and generalizability in segmenting unseen tumors across diverse anatomical regions.

\noindent \textbf{Generalization to Unseen Image Modalities.}
In this setup, we used KiTS23~\cite{heller2023kits21} and four CT tumor segmentation datasets from the MSD~\cite{antonelli2022medical}, including colon, pancreas, lung, and hepatic vessel tumors, to train the models and then tested them on three MRI datasets, including an in-house MRI liver tumor dataset, MSD-Brain~\cite{antonelli2022medical}, and BraTS23~\cite{adewole2023brain}. This experimental setup defines a much more challenging scenario, where models must handle both unseen tumor types and imaging modalities.
The results are shown in~\cref{tab:modality}. %For the in-house MRI liver tumor dataset, %where the models encountered a seen tumor category but in an unseen modality, 
\textbf{DiffuGTS} surpass the competing methods by at least $21.28\%$ in DSC. %For the MSD-Brain~\cite{antonelli2022medical} and BraTS23~\cite{adewole2023brain} datasets, %which presented a more challenging scenario involving both unseen tumor types and modalities, 
%DiffuGTS outperformed the baselines by at least $20.18\%$ in DSC. 
Due to the significant differences between modalities, most competing methods suffer a considerable drop in performance. This is because even the visual features of the same anatomical structures can vary greatly across different modalities. As shown in row four of~\cref{fig:diffuGTS_intro}, our method generates high-quality pseudo-healthy brain images, where the tumor regions significantly differ from the original, leading to a substantial improvement in tumor segmentation performance. At the same time, the text-driven AOVA map accurately captures the potential tumor locations, benefiting from the internal representations of the frozen VAE encoder. These results further support our motivation, showing that effectively leveraging diverse anatomical knowledge from medical foundational diffusion models, and adapting them for zero-shot tumor segmentation through innovative designs, leads to significantly improved cross-modality generalizability and robustness.

\begin{table}
  \Huge
  %\begin{center}
  \centering
  \resizebox{\linewidth}{!}
  {
  \begin{tabular}{@{}c|c|c|c|c@{}}
    \hline
    \diagbox{Efficiency}{Method} & \textbf{DiffuGTS}  & ZePT~\cite{jiang2024zept} & DDPM-MAD~\cite{wolleb2022diffusion2} & THOR~\cite{bercea2024diffusion}\\
    \hline
     Trainable Params        & 284.96M & 745.94M &   733.27M   &  783.42M      \\
     \hline
     FLOPs                   & 12876.48G      & 3886.95G &   6895.53G   &  7143.30G    \\  
    \hline
  \end{tabular}
  }
  %\end{center}
  \caption{Computational cost comparison between \textbf{DiffuGTS} and some competting methods. The FLOPs is computed based on input with spatial size $128\times 128 \times 128$ on the same A100 GPU. 
  }
  \vspace{-10pt}
  \label{tab:Efficiency}
\end{table}

\noindent \textbf{Computation Efficiency.} As shown in~\cref{tab:Efficiency}, the number of trainable parameters in \textbf{DiffuGTS} is significantly smaller compared to traditional encoder-decoder-based methods, as it utilizes internal features from MAISI, eliminating the need to train a large image encoder and decoder. However, using MAISI for feature extraction and generation results in higher computational costs in terms of FLOPs compared to other methods. This substantial computational overhead is a current limitation of foundational models. Efficient knowledge distillation algorithms present a promising solution, enabling the distillation and reuse of knowledge from foundational models while reducing computational costs. Nonetheless, this approach lies beyond the scope of this study and is reserved for future work.
%The total number of trainable parameters in FDM-GTS is provided in~\cref{tab:Efficiency}.

\noindent \textbf{Qualitative Analysis.} 
%AOVA maps for seen and unseen categories
\cref{fig:QR} shows the qualitative results (leave-one-out setting) and demonstrates the merits of \textbf{DiffuGTS}. Most competing methods suffer from segmentation incompleteness-related failures and misclassification of background regions as tumors (false positives). \textbf{DiffuGTS} generates results that are more consistent with the ground truth in comparison with all other competing models. %This success is attributed to the advantages of query-decoupling scheme and self-prompted learning strategy which enables the advanced queries to learn universal and generalizable features related to pathological changes. 

\subsection{Ablation Study and Discussions}
Ablation studies were conducted by training on the KiTS23 and four CT tumor datasets of MSD, including colon, pancreas, lung, and hepatic vessel tumors, followed by testing on the MSD liver and brain tumor datasets to evaluate generalization to unseen tumors and modalities.

\noindent \textbf{Significance of Leveraging Frozen Internal Representations of MAISI with Adapters.}
We examine the contribution of adapting the internal representations from MAISI's VAE encoder $V_E$ for GTS. 
We experiment with three alternatives: (1) training with an nnUNet~\cite{isensee2021nnu} image encoder from scratch, (2) directly using internal features from frozen $V_E$ without adapters, and (3) fine-tuning $V_E$ instead of adopting adapters. The results are reported in~\cref{tab:internalMAISI}. All these alternatives lead to significant performance degradation. Training with nnUNet's image encoder from scratch misses out on the rich anatomical knowledge embedded in $V_E$. Directly using internal features from $V_E$ without an adapter is also suboptimal, as these features are optimized for a generative task, creating a gap with the requirements of semantic segmentation. Fine-tuning $V_E$ without adapters results in the network overfitting to seen categories, leading to the loss of the original knowledge gained from diffusion training. In contrast, our strategy employs an adapter that repurposes $V_E$ for the GTS task while preserving its learned knowledge by keeping the parameters frozen.

\begin{table}
\Huge
  \centering
  %\begin{center}
  \resizebox{\linewidth}{!}
  {
  \begin{tabular}{@{}l|cc|cc@{}}
    \toprule
    %\toprule
    \multicolumn{1}{l|}{\multirow{2}{*}{Method}} & \multicolumn{2}{c|}{MSD Liver Tumor} & \multicolumn{2}{c}{MSD Brain Tumor}\\
    \cline{2-3}\cline{4-5}
    & DSC$\uparrow$  & NSD$\uparrow$ & DSC$\uparrow$ & NSD$\uparrow$   \\
     %\hline
     %ZePT~\cite{jiang2024zept}       & 94.53                    & 80.11  & 72.46        & 58.93               \\ 
     %ZePT~\cite{jiang2024zept}$_{Adapter}$ (MAISI $V_E$)     & 96.25                  & 81.00
     %& 78.01 & 44.35  \\
    \hline
     DiffuGTS$_{Scratch}$ (nnUNet Encoder)       & 60.40                    & 70.37
     & 32.18      & 34.49            \\
     DiffuGTS$_{w/o \textbf{ } Adapter}$ (MAISI $V_E$)         & 60.89                    & 70.75  
     & 33.92        & 35.90             \\ 
     DiffuGTS$_{Fine-tuning}$ (MAISI $V_E$)      & 61.18                    & 71.59  
     & 34.76        & 36.65             \\ 
     \hline
     DiffuGTS$_{Adapter}$ (MAISI $V_E$)               &  \textbf{63.23} &  \textbf{73.58}  
     & \textbf{47.51}  &\textbf{49.75}    \\
    \bottomrule
  \end{tabular}
  }
  %\end{center}
  \caption{Ablation study of leveraging frozen internal representations of MAISI with adapters. ``w/o'' denotes ``without''.}
  \vspace{-10pt}
  \label{tab:internalMAISI}
\end{table}

\noindent \textbf{Why Does DiffuGTS Generalize Well?}
In~\cref{tab:NC}, we examine the contribution of two key components of \textbf{DiffuGTS}. (1) Importance of AOVA maps. We replace the AOVA maps with a query-based Mask2Former~\citep{cheng2022masked} backbone, which is widely used in open-vocabulary~\cite{liang2023open,ghiasi2022scaling,qin2023freeseg} or zero-shot~\cite{ding2022decoupling,jiang2024zept,jiang2024unleashing} segmentation methods. This leads to a significant performance drop of $2.02\%$ in DSC and $1.95\%$ in NSD for unseen liver tumor segmentation, as well as a drop of $10.94\%$ in DSC and $11.26\%$ in NSD for unseen brain tumor segmentation in MRI. 
Mask2Former updates its object queries based on the features of the training images. This additional parameter optimization leads to overfitting of the object queries to the seen categories, thereby degrading performance on unseen tumors.
In contrast, our AOVA directly uses text embeddings to establish cross-modal correlations, thereby better leveraging the frozen VAE encoder’s generalization capabilities. %These results confirm the superiority of AOVA maps for cross-modal, text-driven anomaly segmentation mask prediction. 
(2) Effectiveness of mask refinement (MR) with frozen MAISI. %We experiment with two alternative operations 
We demonstrate that utilizing the MAISI model for mask refinement (AOVA + MR) significantly improves segmentation performance compared to relying solely on text-driven anomaly maps for mask predictions (AOVA only). This supports our motivation that employing frozen foundation diffusion models for mask refinement improves the quality of text-driven segmentation masks.
Furthermore, removing the mask refinement process with diffusion models leads to a more pronounced performance decline than substituting AOVA with Mask2Former. This underscores the importance of using mask refinement to enhance the model's zero-shot generalization ability.

\begin{figure}
  \centering
%  %\fbox{\rule{0pt}{2in} \rule{0.9\linewidth}{0pt}}
  \includegraphics[width=\linewidth]{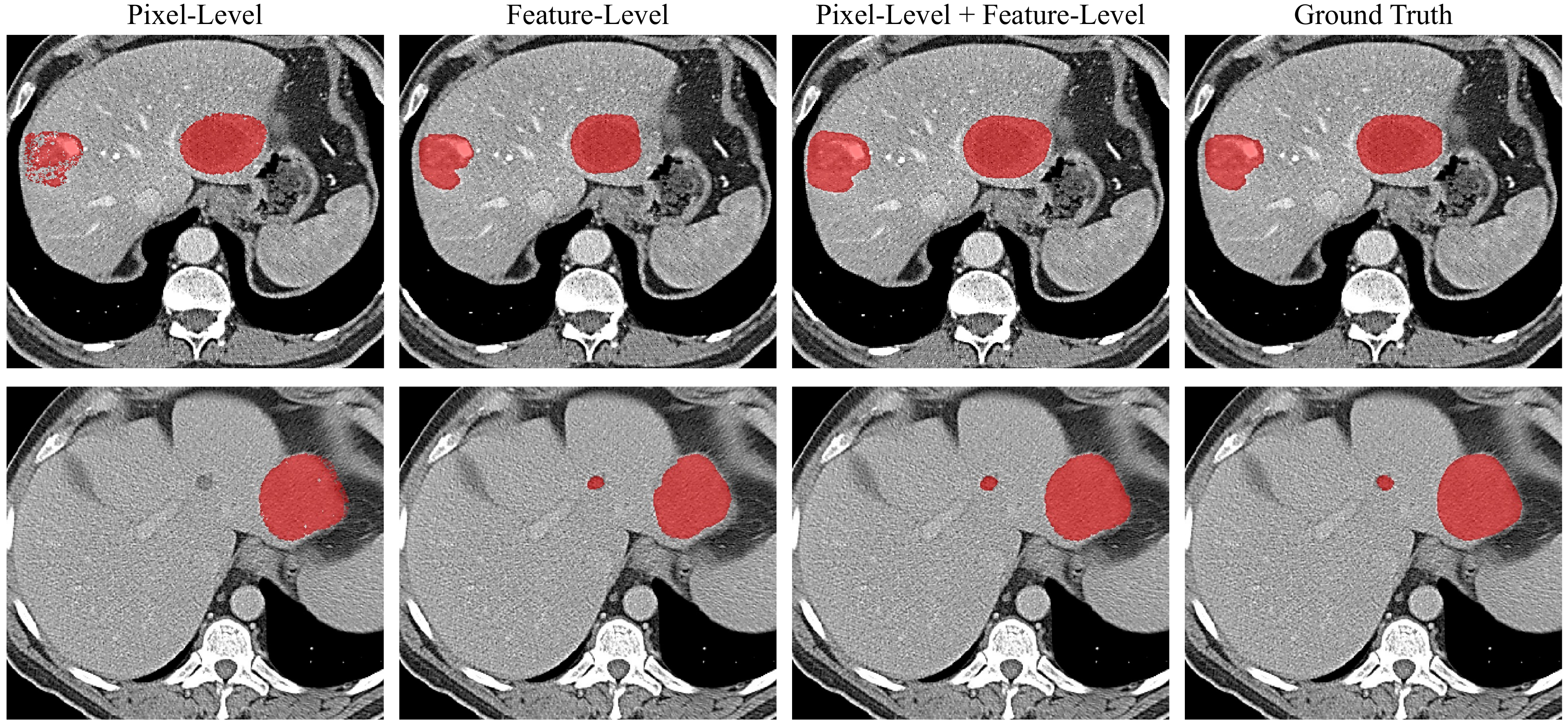}
  \caption{Visualization illustrating how utilizing pixel-level and feature-level residual learning improves performance. %We present several prediction cases generated using both text embeddings and mask tokens, text embeddings only (TE), and mask tokens only (MT).
  }
  \vspace{-5pt}
  \label{fig:PH}
\end{figure}

\noindent \textbf{Effectiveness of pixel-level and feature-level Residual Learning.} In~\cref{tab:pixelvsfeature}, we compare the effect of using pixel-level residual maps $P_r$, feature-level residual maps $F_r$, or a combination of both in calculating the final segmentation results. It can be observed that combining pixel-level and feature-level residual maps leads to a better segmentation performance. \cref{fig:PH} provides a visual comparison, illustrating the enhancement achieved by incorporating feature-level residual learning over relying solely on pixel-level residuals like previous methods~\cite{wolleb2022diffusion2,bercea2024diffusion}.

\begin{table}
  \Huge
  \centering
  %\begin{center}
  \resizebox{\linewidth}{!}
  {
  \begin{tabular}{@{}l|cc|cc@{}}
    \toprule
    %\toprule
    \multicolumn{1}{l|}{\multirow{2}{*}{Method}} & \multicolumn{2}{c|}{MSD Liver Tumor} & \multicolumn{2}{c}{MSD Brain Tumor}\\
    \cline{2-3}\cline{4-5}
    & DSC$\uparrow$  & NSD$\uparrow$ & DSC$\uparrow$ & NSD$\uparrow$   \\
    \hline
     DiffuGTS (Mask2Former~\cite{cheng2022masked} + MR)      & 61.21 & 71.63        & 36.57     & 38.49                     \\
     DiffuGTS (AOVA only)      & 59.94                    & 70.11                   & 20.14     & 22.90                     \\
     DiffuGTS (AOVA + MR)             &  \textbf{63.23} &  \textbf{73.58}           & \textbf{47.51}  &\textbf{49.75}       \\
    \bottomrule
  \end{tabular}
  }
  %\end{center}
  \caption{Ablation study of the proposed AOVA maps and mask refinement process with diffusion models.}
  \vspace{-5pt}
  \label{tab:NC}
\end{table}

%~\cref{tab:selfprompted} shows the investigation on visual prompts.

\begin{table}
  \centering
  %\begin{center}
  \resizebox{0.9\linewidth}{!}
  {
  \begin{tabular}{@{}l|cc|cc@{}}
    \toprule
    %\toprule
    \multicolumn{1}{l|}{\multirow{2}{*}{Method}} & \multicolumn{2}{c|}{MSD Liver Tumor} & \multicolumn{2}{c}{MSD Brain Tumor}\\
    \cline{2-3}\cline{4-5}
    & DSC$\uparrow$  & NSD$\uparrow$ & DSC$\uparrow$ & NSD$\uparrow$   \\
    \hline
     DiffuGTS ($P_r$)       & 61.85                    & 71.82           & 38.79      & 40.66            \\
     DiffuGTS ($F_r$)      & 61.94                    & 72.05            & 41.14        & 43.27             \\  
     DiffuGTS ($P_r + F_r$)  &  \textbf{63.23}     &  \textbf{73.58}      & \textbf{47.51}  &\textbf{49.75}   \\
    \bottomrule
  \end{tabular}
  }
  %\end{center}
  \caption{Ablation study of the proposed pixel-level and feature-level residual learning.}
  \vspace{-5pt}
  \label{tab:pixelvsfeature}
\end{table}

%\noindent \textbf{Limitations and Future Work.}

%Finally, we would like to discuss the limitations of the current work as well as the future directions to further improve it. 
%In more complex tasks...
%We hope our work can shed some light on designing ... tailored for ... tasks.
%\noindent \textbf{Discussions about Limitations.} 
\section{Conclusions}
\label{conclusion}
In this work, we unlock advanced generalizable tumor segmentation from frozen medical foundation diffusion models by introducing a novel framework named \textbf{DiffuGTS}. \textbf{DiffuGTS} employs a series of carefully designed strategies to initially construct anomaly-aware open-vocabulary attention maps for tumor detection. It then utilizes a frozen medical foundational diffusion model for further anomaly mask refinement, demonstrating superior zero-shot tumor segmentation capabilities across various anatomical regions and imaging modalities. %Looking ahead, with anticipated advancements in medical foundation diffusion models, 
We hope our method provides insights into efficiently leveraging foundation diffusion models for zero-shot tumor segmentation tasks.
\par
\section*{Acknowledgments}
This work is funded by the National Key R\&D Program of China under grants No.2022ZD0160700 and is supported by Shanghai Artificial Intelligence Laboratory.
{
    \small
    \bibliographystyle{ieeenat_fullname}
    \bibliography{main}
}

% WARNING: do not forget to delete the supplementary pages from your submission 
\clearpage
\setcounter{page}{1}
\maketitlesupplementary

\appendix
\section{Dataset Details}
\label{sec:Dataset}
Our study utilizes datasets encompassing tumors across $7$ diseases and $6$ organs, derived from both public and private sources. We summarize all the datasets in Table~\ref{tab:datasets}.

\subsection{Public Datasets}
\textbf{KiTS23}. This dataset is from the Kidney and Kidney Tumor Segmentation Challenge~\cite{heller2023kits21}, which provides $489$ cases of data with annotations for the segmentation of kidneys, renal tumors, and cysts. 

\noindent \textbf{MSD}. The datasets of liver tumor, pancreas tumor, colon tumor, lung tumor, and brain tumor are part of the Medical Segmentation Decathlon (MSD)~\cite{antonelli2022medical}, providing annotated datasets for various tumors.

\noindent \textbf{BraTS23}. This dataset is part of the RSNA-ASNR-MICCAI BraTS 2023 Challenge~\cite{adewole2023brain}, comprising $1,251$ multi-institutional, clinically-acquired multi-parametric MRI (mpMRI) scans of glioma. The ground truth annotations include sub-regions used for evaluating the 'enhancing tumor' (ET), 'non-enhancing tumor core' (NETC), and 'surrounding non-enhancing FLAIR hyperintensity' (SNFH). In this study, we adopt the 'whole tumor' setting, which describes the complete extent of the disease, for segmentation evaluation.

\noindent \textbf{TotalSegmentator}. TotalSegmentator~\cite{wasserthal2023totalsegmentator} collects $1024$ CT scans randomly sampled from PACS over the timespan of the last 10 years. The dataset contains CT images with different sequences
(native, arterial, portal venous, late phase, dual-energy), with and without contrast agent, with different bulb voltages,
with different slice thicknesses and resolution and with different kernels (soft tissue kernel, bone kernel). A total of 404 patients showed no signs of pathology, and their data are used in our study as healthy samples for anomaly detection training.

\subsection{Private Datasets}
This dataset comprises a large number of
high-resolution T2-weighted 3D MRI images from a
total of $400$ patients. We acquired one volume from each patient. The segmentation ground truths are provided for each volume in the dataset. All liver tumors and surrounding normal tissues were segmented manually by one radiologist and confirmed by another. During the annotation phase, the radiologists are also provided with the corresponding post-surgery pathological report to narrow down the search area for the tumors. %If the two radiologists disagreed, a third radiologist with more than 25 years of experience would reevaluate the delineated boundaries as the final results.
All the MRI scans share the same in-plane dimension of $512 \times 512$, and the dimension along the z-axis ranges from $85$ to $225$, with a median of $155$. The in-plane spacing ranges from $0.45\times0.45$ to $0.62\times0.62$ mm, with a median of $0.53 \times 0.53$ mm, and the z-axis spacing is from $3.0$ to $5.5$ mm, with a median of $4.2$ mm. 

\subsection{Preprocessing}
We adopt similar data processing strategies as used in MAISI~\cite{guo2024maisi}. For CT images, the intensities are clipped to a Hounsfield Unit (HU) range of $-1000$ to $1000$ and normalized to a range of $[0,1]$. For MR images, intensities are normalized such that the $0$th to $99.5$th percentile values are scaled to the range $[0,1]$. Intensity augmentations for MR images include random bias field, random Gibbs noise, random contrast adjustment, and random histogram shifts. Both CT and MR images undergo spatial augmentations, such as random flipping, random rotation, random intensity scaling, and random intensity shifting.

\begin{table*}
 \centering
  \resizebox{0.9\linewidth}{!}
  {
    \begin{tabular}{c|c|c|c|c}
	\toprule
    \textbf{Data Source} & \textbf{Modality} &\textbf{Dataset Name} & \textbf{Segmentation  Targets} & \textbf{Number of scans} \\ 
      \hline
\multirow{9}{*}{Public} & \multirow{6}{*}{CT} 
&KiTS23~\cite{heller2023kits21} & Kidney Tumor, Kidney Cyst & 489 \\
&&MSD-Colon~\cite{antonelli2022medical}  & Colon Tumor & 126 \\
&&MSD-Liver~\cite{antonelli2022medical}  & Liver Tumor & 131 \\
&&MSD-Hepatic Vessel~\cite{antonelli2022medical}  & Hepatic Vessel Tumor & 303 \\
&&MSD-Lung~\cite{antonelli2022medical}  & Lung Tumor & 64 \\
&&MSD-Pancreas~\cite{antonelli2022medical} & Pancreas Tumor & 281 \\
\cline{3-5}
&&TotalSegmentator~\cite{wasserthal2023totalsegmentator} & Kidney, Lung, Pancreas, Colon, Liver, Brain & 404 \\
\cline{2-5}
&\multirow{2}{*}{MRI}
&MSD-Brain~\cite{antonelli2022medical} & Gliomas & 484 \\
&&BraTS23~\cite{adewole2023brain} & Gliomas & 1251 \\
    \hline
Private & MRI & in-house dataset & Liver Tumor & 400 \\
    \bottomrule
		\end{tabular}
        }
\caption{Details of Datasets.}
 \label{tab:datasets}
\end{table*}

\section{More Qualitative Analysis.}
\begin{figure*}[htbp]
  \centering
  \includegraphics[width=\linewidth]{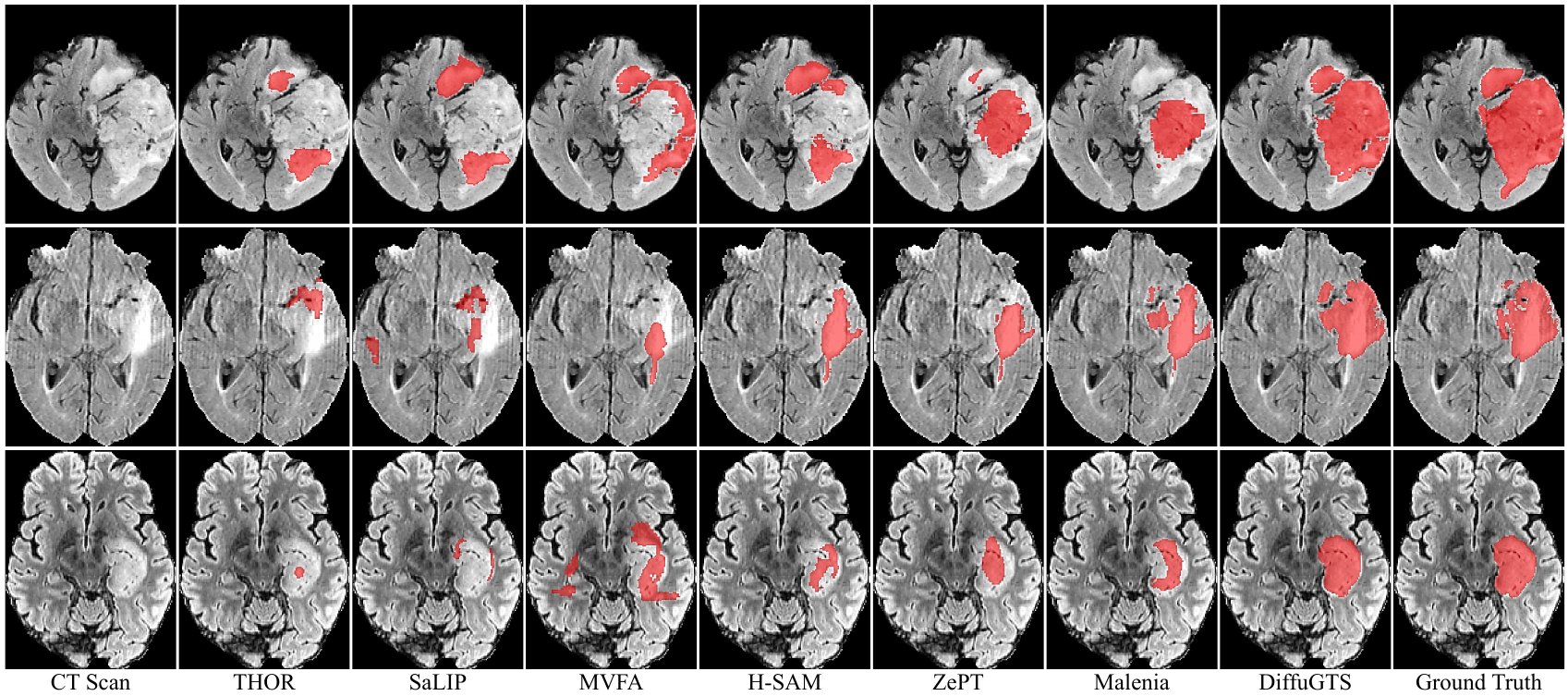}
  \caption{Qualitative visualizations of zero-shot segmentation results on BraTS23~\cite{adewole2023brain}.}
  %\vspace{-10pt}
  \label{fig:brain}
\end{figure*}

For qualitative analysis on BraTS23~\cite{adewole2023brain}, we present visualizations of segmentation results in~\cref{fig:brain}. This shows that our approach achieves much better zero-shot cross-modality generalization performance compared with other competing methods.

\section{Additional Ablation Experiments}
In line with the ablation study setting in the main paper, where the model is trained on the KiTS23 dataset and four CT tumor datasets from MSD, including colon, pancreas, lung, and hepatic vessel tumors, followed by testing on the MSD liver and brain tumor datasets to evaluate generalization to unseen tumors and modalities, we conduct extensive ablation studies for further evaluation.

\noindent \textbf{Significance of Multi-scale Feature Aggregation}
We aggregate cross-attention matrices between text-attribution keys and pixel queries across three feature levels to generate the AOVA maps. We conduct ablation experiments to examine the efficacy of utilizing multi-scale image features from the MAISI VAE encoder. The outcomes, elucidated~\cref{tab:Multiscale}, provide a comprehensive understanding of the performance gains achieved through multi-scale feature aggregation for constructing AOVA maps, compared to using single-level image features.
% \begin{table}
%     \centering
%       %\resizebox{\linewidth}{!}
%   %{    
% \begin{tabular}{@{}ccc|cc|cc@{}}
%     \toprule
%     \multicolumn{3}{c}{Levels} & \multicolumn{2}{|c}{MSD-Liver Tumor} & \multicolumn{2}{|c}{BraTS23} \\
%      \cmidrule(lr){1-3}\cmidrule(lr){4-5}\cmidrule(lr){6-7}
%     Level 1 & Level 2 & Level 3 & DSC$\uparrow$ & NSD$\uparrow$ & DSC$\uparrow$ & NSD$\uparrow$ \\
%     \midrule
%     \checkmark  & &                                                     &38.96 & 54.79        &47.24   &53.68     \\
%     & \checkmark        &                                               &40.27 &56.34         &49.09   &55.14    \\
%     \checkmark  &\checkmark   &                                         &41.22 & 57.15         &52.58   &58.50     \\
%      \rowcolor{red!8}\checkmark &  & \checkmark                         &40.50 & 56.61    &51.77   &57.83        \\
%       \rowcolor{red!8}& \checkmark & \checkmark                         &42.44 & 58.59         &52.87   &58.63 \\
%     \rowcolor{red!8}\checkmark & \checkmark  & \checkmark               &\textbf{43.30} &\textbf{59.63}  &\textbf{54.96} &\textbf{60.60}     \\ 
%     \bottomrule
%   \end{tabular}
%   %}
%     \caption{Ablation study of multi-scale feature aggregation for constructing AOVA maps. The DSC and NSD are reported. The best result is in light blue.}
%      \label{tab:Multiscale}
% \end{table}

\begin{table}
    \centering
      \resizebox{0.9\linewidth}{!}
  {    
\begin{tabular}{@{}c|cc|cc@{}}
    \toprule
    \multirow{2}{*}{Feature Levels} & \multicolumn{2}{|c}{MSD Liver Tumor} & \multicolumn{2}{|c}{MSD Brain Tumor} \\
    \cline{2-3}\cline{4-5}
    & DSC$\uparrow$ & NSD$\uparrow$ & DSC$\uparrow$ & NSD$\uparrow$ \\
    \midrule
    Level 1 &62.07 & 72.16        &43.40   &45.33     \\
    Level 2 &62.28 &72.49         &44.92   &46.84    \\
    Level 3 &62.13 & 72.35         &43.88   &45.96     \\
    \rowcolor{blue!8}  Aggregation  & \textbf{63.23} &  \textbf{73.58}  & \textbf{47.51}  &\textbf{49.75}       \\
    \bottomrule
  \end{tabular}
  }
    \caption{Ablation study of multi-scale feature aggregation for constructing AOVA maps. The DSC and NSD are reported. The best result is in light blue.}
     \label{tab:Multiscale}
\end{table}

\begin{figure*}
  \centering
  \includegraphics[width=0.9\linewidth]{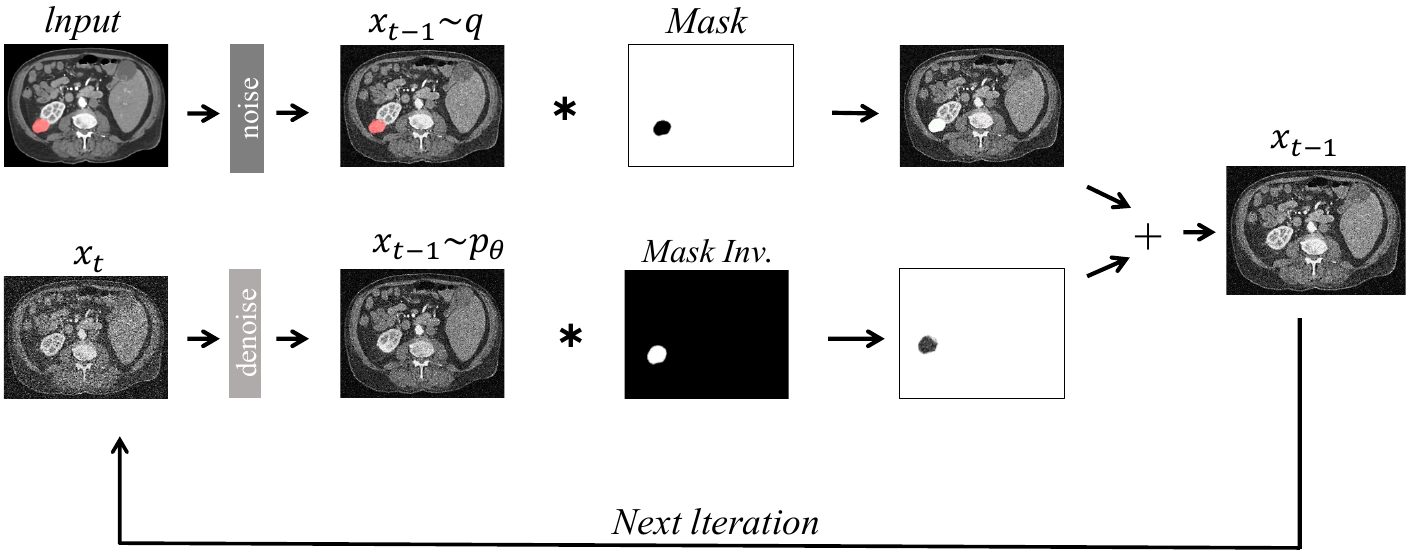}
  \caption{The illustration of the one-step reverse process of the inpainting strategy.}
  %\vspace{-10pt}
  \label{fig:reverse}
\end{figure*}

\noindent \textbf{Effectiveness of Latent Space Inpainting.} 
We demonstrate the impact of using versus not using training-free latent space inpainting (LSI) strategy when generating pseudo-healthy equivalents in~\cref{tab:LSI}. Directly applying MAISI for the generation leads to substantial changes in the healthy regions of the target organ (also shown in~\cref{fig:Latent}), which subsequently decreases segmentation performance. In contrast, our strategy effectively preserves details in the organ that are unaffected by the disease, underscoring the importance of modifying the generation process of the original MAISI~\cite{guo2024maisi} through latent space inpainting strategy. Additionally, this approach is entirely training-free, avoiding the computational costs associated with retraining or fine-tuning a foundational diffusion model. The illustration of the one-step reverse process of the inpainting strategy is shown in~\cref{fig:reverse}.

\begin{table}
    \centering
    \resizebox{\linewidth}{!}{
\begin{tabular}{@{}c|cc|cc@{}}
    \toprule
    \multirow{2}{*}{Method} & \multicolumn{2}{|c}{MSD Liver Tumor} & \multicolumn{2}{|c}{MSD Brain Tumor} \\
    \cline{2-3}\cline{4-5}
    & DSC$\uparrow$ & NSD$\uparrow$ & DSC$\uparrow$ & NSD$\uparrow$ \\
    \midrule
    DiffuGTS$_{MAISI}$ &60.36 & 70.31        &30.55   &32.74     \\
    \rowcolor{blue!8}  DiffuGTS$_{MAISI}$ + LSI  & \textbf{63.23} &  \textbf{73.58}  & \textbf{47.51}  &\textbf{49.75}       \\
    \bottomrule
    \end{tabular}
    }
    \caption{Ablation study on leveraging the latent space inpainting (LSI) strategy to generate pseudo-healthy equivalents, compared to directly using MAISI for generation. The DSC and NSD metrics are reported.}
     \label{tab:LSI}
\end{table}

\noindent \textbf{Is the improvement solely attributed to the MAISI?} 
To leverage the capabilities of the medical foundational diffusion model, we introduce a series of sophisticated designs and demonstrated their effectiveness through ablation studies. Additionally, we conduct further experiments to show that the performance improvements are not merely due to utilizing the medical foundational diffusion model, but largely stemmed from our innovative designs. To validate this, we apply the MAISI VAE encoder to some existing methods and use MAISI to refine the masks generated by these methods.

The comparison results are shown in~\cref{tab:faircompare}. We observe that using the VAE encoder from MAISI for image feature extraction and employing MAISI's generative capability to further refine the masks enhances the performance of existing methods. This supports our motivation for leveraging foundational diffusion models for advanced zero-shot tumor segmentation. Furthermore, even when existing methods benefit from MAISI's capabilities and knowledge, DiffuGTS consistently outperforms them. \textbf{This demonstrates that the improvement in zero-shot generalization performance is not solely due to the foundational diffusion model, but also attributed to our innovative designs}, which effectively unleash the potential of utilizing foundation diffusion model for generalizable tumor segmentation.

\begin{table}
    \centering
    \resizebox{\linewidth}{!}{
\begin{tabular}{@{}c|cc|cc@{}}
    \toprule
    \multirow{2}{*}{Method} & \multicolumn{2}{|c}{MSD Liver Tumor} & \multicolumn{2}{|c}{MSD Brain Tumor} \\
    \cline{2-3}\cline{4-5}
    & DSC$\uparrow$ & NSD$\uparrow$ & DSC$\uparrow$ & NSD$\uparrow$ \\
    \midrule
    ZePT~\cite{jiang2024zept}                               &59.16 & 68.72        &19.54   &22.02     \\
    Malenia~\cite{jiang2024unleashing}                      &59.83 & 70.08        &19.83   &22.58     \\
    ZePT~\cite{jiang2024zept} + MAISI~\cite{guo2024maisi}   &60.16 & 70.14        &27.21   &29.53     \\
    Malenia~\cite{jiang2024unleashing} + MAISI~\cite{guo2024maisi}  &60.28 & 70.22     &27.86   &29.94     \\
    \rowcolor{blue!8}  \textbf{DiffuGTS}  & \textbf{63.23} &  \textbf{73.58}  & \textbf{47.51}  &\textbf{49.75}       \\
    \bottomrule
    \end{tabular}
    }
    \caption{Comparisons between \textbf{DiffuGTS} and existing methods combined with MAISI.}
     \label{tab:faircompare}
\end{table}

\section{Model Robustness Analysis}
\begin{figure}
  \centering
  \includegraphics[width=\linewidth]{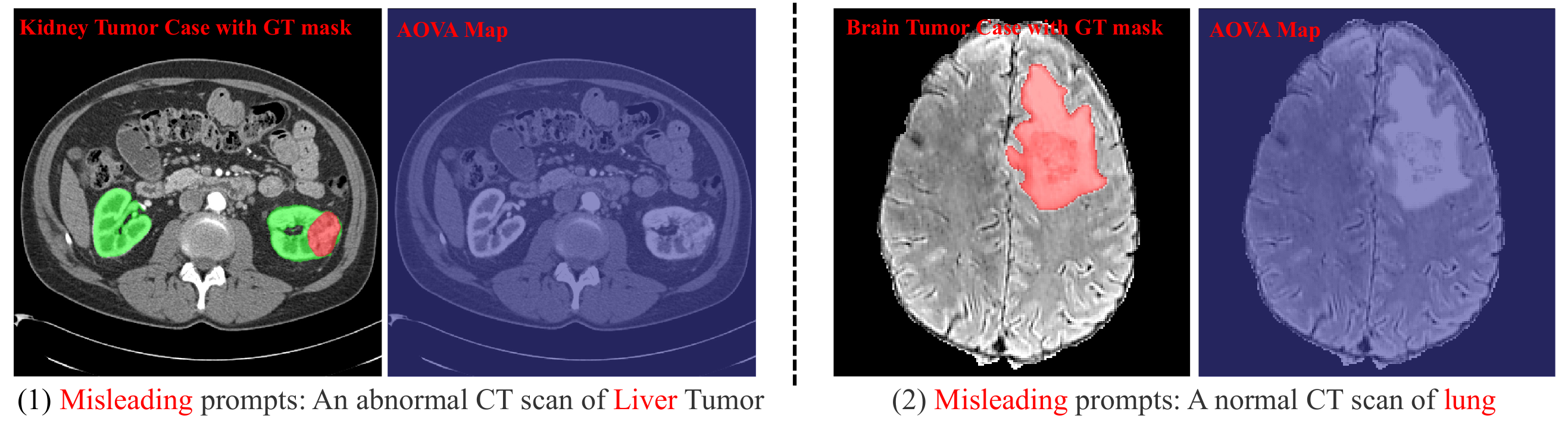}
  \caption{How the model processes misleading text prompts.}
  \label{fig:rebuttal_vis}
\end{figure}

In~\cref{fig:rebuttal_vis}, we show how our model handles misleading prompts: (1) a disease that is not present, and (2) using a lung-related prompt on a brain scan. The AOVA maps generated by these prompts exhibit no strong activation, indicating that the model recognizes that none of the image content is relevant to the text prompts and therefore does not predict any foreground mask. This further demonstrates that our model has effectively learned the correlations between visual features and textual descriptions, achieving a genuine understanding of anatomical structures.

\section{Explanation of ``Pseudo-Healthy'' Images}
We would like to further clarify that the generated pseudo-healthy images are not actual healthy images. Similar to many diffusion-based medical anomaly detection methods~\cite{wolleb2022diffusion2,bercea2024diffusion}, the primary purpose of generating these pseudo-healthy images is to segment tumors by highlighting the differences between the original image and the generated image. Ideally, the generated image should exhibit significant changes in the tumor region while preserving the non-tumor areas of the original image, regardless of whether those areas are healthy. Thus, the term "ideal" here specifically refers to tumor segmentation, rather than implying the generation of a completely healthy image. In other words, the generated pseudo-healthy images only need to preserve the non-tumor areas of the original image while significantly altering the tumor regions, rather than striving to create a fully healthy image. Additionally, whether non-tumor regions of an organ with tumors can still be considered "healthy" is a broader discussion beyond the technical scope of this paper. To prevent any misunderstandings, we refer to these generated images as "pseudo-healthy" images.

%\section{Segmentation of Seen Organs and Tumors.}
%Table presents the quantitative results comparing \textbf{DiffuGTS} with existing state-of-the-art zero-shot tumor segmentation methods. Our method consistently outperforms the baselines in terms of Dice Similarity Coefficient (DSC) and Normalized Surface Distance (NSD), demonstrating superior segmentation accuracy across all evaluated tumor categories.
%Our method's success in zero-shot settings underscores its potential for practical clinical applications where annotated data for certain tumor types may be scarce or unavailable. The ability to generalize to unseen categories without additional training highlights the effectiveness of exploiting foundational diffusion models and open-vocabulary attention mechanisms.

\section{Analysis of Potential Data Leakage}
We used MSD, KiTS23, BraTS23, and an in-house liver tumor dataset for evaluation. Among these, only the MSD overlaps with the dataset used during MAISI’s training. %the other datasets were not involved in training the MAISI foundational model. 
A key concern is whether the MAISI framework inadvertently introduces label information leakage that could compromise the model's training independence. 
In this section, we conduct a rigorous analysis of this critical issue.
Apparently, the performance improvement of our framework is not exclusively derived from the MAISI integration. As validated in~\cref{tab:faircompare}, the principal performance improvement mainly stems from our innovative designs.
Furthermore, we clarify that our framework does not leak any label information from MAISI related to the MSD dataset into downstream testing. First, we use the internal features of the MAISI VAE encoder. 
%the training objective of the MAISI VAE encoder and decoder 
The MAISI VAE encoder and decoder were trained on the volume reconstruction task, which only involved image data and did not use any mask annotations. 
%Therefore, there is no risk of data leakage when we use MAISI VAE encoder's internal features to train the AOVA maps.
Therefore, using the MAISI VAE encoder's internal features to train the AOVA maps poses no risk of data leakage.
%Second, regarding the Diffusion Model in MAISI, its training objective on MSD was to generate tumors given a tumor mask.
%In contrast, we also provide a tumor mask but force the model to generate pseudo-healthy organs, establishing an out-of-distribution inference scenario.
Second, the diffusion model in MAISI is trained on the MSD dataset to synthesize tumors explicitly conditioned on a tumor mask via ControlNet. In contrast, our method utilizes a coarse tumor mask implicitly through a repaint mechanism, forcing the model to generate pseudo-healthy organs instead of tumors. This fundamental divergence in conditioning strategies shifts the MAISI's inference paradigm from %while MAISI operates within 
an in-distribution scenario (tumor generation aligned with MAISI's training data) to %our approach establishes 
an out-of-distribution scenario (synthesizing healthy anatomy from anomalous inputs).
This approach essentially prevents the diffusion model from utilizing any memorized label information.
%If data leakage were to occur, the model would reconstruct the tumor, not the pseudo-healthy organ we aim for since the masked image input serves as a strong hint for tumor generation. 
If data leakage were to occur, the model would generate the tumor rather than the pseudo-healthy organ we intend. %since the tumor mask serves as a strong cue for tumor generation.
Additionally, generating pseudo-healthy organs on MSD is not involved in MAISI's training. These support the claim that our framework does not leak any label information from MAISI related to the MSD dataset into downstream segmentation testing. 
Moreover, the superior performance of \textbf{DiffuGTS} on KiTS23, BraTS23, and our in-house liver tumor dataset—all excluded from the MAISI foundation model's training data—demonstrates the generalizability and robustness of our proposed strategies.

\section{Limitations and Future Work}
Our method, through carefully crafted innovative designs, has unleashed the potential of medical foundational diffusion models for advanced zero-shot 3D tumor segmentation. However, it remains constrained by the capabilities of the underlying medical foundational diffusion model. As the MAISI VAE is designed as a foundational model for 3D CT and MRI, our research is similarly limited to these imaging modalities, leaving other modalities, such as 2D X-ray, unaddressed. In future research, we aim to explore zero-shot multimodal models that encompass a broader range of imaging modalities and clinical scenarios. Furthermore, as medical foundational diffusion models continue to evolve, our method stands to benefit from these advancements, with the potential for further enhancement in performance.

\end{document}